\pdfoutput=1  

\documentclass[10pt,twocolumn,letterpaper]{article}

\usepackage[letterpaper,margin=0.75in,columnsep=0.25in]{geometry}

\usepackage[T1]{fontenc}
\usepackage[utf8]{inputenc}

\usepackage{amsmath}
\usepackage{amssymb}
\usepackage{amsfonts}

\IfFileExists{newtxtext.sty}
  {\usepackage{newtxtext,newtxmath}}
  {\usepackage{mathptmx}}
\usepackage{microtype}

\usepackage{graphicx}
\usepackage{booktabs}
\usepackage{multirow}
\usepackage[table]{xcolor}
\usepackage{caption}
\usepackage{float}
\usepackage{algorithm}
\usepackage{algorithmic}
\usepackage{listings}
\usepackage{soul}
\usepackage{xspace}
\usepackage{etoolbox}

\usepackage[numbers,sort&compress]{natbib}
\usepackage[hyphens]{url}
\usepackage[colorlinks=true,linkcolor=black,citecolor=black,%
            urlcolor=blue,breaklinks=true]{hyperref}

\DeclareCaptionStyle{ruled}{labelfont=normalfont,labelsep=colon,strut=off}
\floatstyle{ruled}
\newfloat{listing}{tb}{lst}{}
\floatname{listing}{Listing}

\newbool{showComments}
\setbool{showComments}{true}
\newcommand{\SysName}{CardioState-JEPA\xspace}

\title{\bf CardioState-JEPA: Delay-Aware Cross-Modal Learning\\
of a Shared Cardiac Representation}

\author{
  Hamza Shafiq\textsuperscript{1}\quad
  Hung Manh Pham\textsuperscript{2}\quad
  Bin Zhu\textsuperscript{2}\quad
  Pan Zhou\textsuperscript{2}\quad
  Jun Hu\textsuperscript{1}\quad
  Aaqib Saeed\textsuperscript{1}\thanks{Corresponding author: \texttt{a.saeed@tue.nl}}
  \\[4pt]
  \normalsize\textsuperscript{1} Eindhoven University of Technology, Netherlands\\
  \normalsize\textsuperscript{2} Singapore Management University, Singapore\\[3pt]
  \small\textbf{Code:} \url{https://github.com/hamzashafiq28/CardioState-Jepa}\quad
  \textbf{Model weights:} \url{https://huggingface.co/hamzashafiq/CardioState-Jepa}
}
\date{}

\begin{document}

\twocolumn[
\begin{@twocolumnfalse}
\maketitle
\begin{abstract}
\noindent
Electrocardiography (ECG), photoplethysmography (PPG), and phonocardiography (PCG) provide complementary views of the same cardiac cycle, yet existing cardiac foundation models are trained for a single sensing modality, leaving the shared physiology across sensors unexploited. We introduce \SysName, a cardiac foundation model to learn a single shared representation jointly across ECG, PPG, and PCG, built on a physiology-aware joint-embedding predictive architecture. The model maps heterogeneous waveforms into a common token space, processes them with a single shared Transformer encoder, and learns by predicting masked latent cardiac states, placing the pretraining target on shared physiology rather than sensor-specific waveform appearance. To handle the temporal offsets between electrical, mechanical, and hemodynamic events, cross-modal prediction uses a learned delay aligner that matches signals at the corresponding cardiac time. Because synchronized multi-sensor recordings are scarce, \SysName first learns within-modality structure from abundant unimodal data and then uses paired data to align modalities in latent cardiac time. Evaluated as a frozen encoder across 25 downstream tasks spanning ECG, PPG, and PCG, our encoder improves average PPG classification by 8.2 AUROC points, PCG murmur detection by 18.8 AUROC points, and ECG classification by 15.5 AUROC points over the best self-supervised signal baseline and matches or exceeds cardiac models trained with privileged clinical text or supervised labels on several ECG benchmarks. These results establish that heterogeneous cardiac signals can mutually supervise a single foundation model of cardiac physiology.
\end{abstract}
\vspace{1.5em}
\end{@twocolumnfalse}
]

\section{Introduction}
\begin{figure}[t]
\centering
\includegraphics[width=\columnwidth]{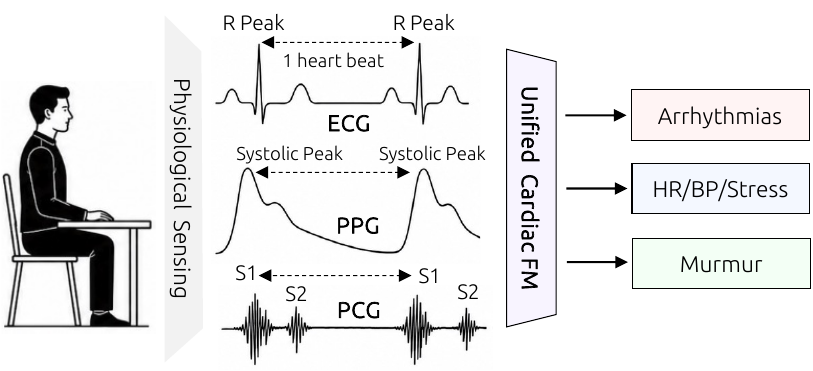}
\caption{Motivation of \SysName. ECG, PPG, and PCG observe the same cardiac cycle at different physiological times, motivating a shared cardiac representation.}
\label{fig:motivation}
\end{figure}

Cardiac signals are a common and important source of information in clinical diagnosis, long-term monitoring, wearable sensing, and everyday health assessment~\cite{lin2021wearable}. Among the many ways to observe cardiac activity, electrocardiography (ECG), photoplethysmography (PPG), and phonocardiography (PCG) remain especially important, as they capture electrical, hemodynamic, and acoustic aspects of the heart, respectively. These modalities are increasingly used across both clinical and daily settings, from arrhythmia analysis and heart sound screening to wearable monitoring of pulse, blood pressure, and physiological stress~\cite{wesad,mimicppg2026,PhysioNet-circor-heart-sound-1.0.3}. As shown in Figure~\ref{fig:motivation}, the three modalities are complementary observations of shared cardiac activity but provide different views of one underlying physiological process. Electrical activation initiates contraction, valve motion then produces the heart sounds, and blood ejection appears later as a peripheral pulse, so the three modalities observe the same beat at different physiological times.

Despite this physiological connection, machine learning for cardiac signals has largely progressed along separate modality-specific paths. Recent advances in large-scale representation learning and foundation models have substantially improved performance for individual cardiac sensors, with ECG models supporting rhythm, conduction, and arrhythmia analysis~\cite{mckeen2025ecgfm,dbeta,liu2024zero}, PPG models enabling wearable tasks such as heart rate estimation, blood pressure prediction, and vascular state assessment~\cite{papagei2025}, and PCG models targeting acoustic signatures such as murmurs and valvular abnormalities~\cite{reyna2023physionet}. However, these models are still typically trained and deployed within a single sensing modality, so structure learned from one view of cardiac physiology is not explicitly used to improve another. A unified cardiac representation model should instead exploit the fact that electrical, hemodynamic, and acoustic signals are different observations of the same underlying cardiac process.

However, building such a model requires more than applying standard multimodal learning. ECG, PPG, and PCG are coupled by physiology but are shifted in time: PCG follows electrical activation via electromechanical coupling, while PPG arrives later via pulse transit. At the same time, the same cardiac event can appear as a sharp electrical deflection, an acoustic burst, or a smooth peripheral pulse wave. Consequently, naive timestamp alignment can compare different phases of the same beat, while raw cross-signal reconstruction can encourage sensor-specific shortcuts. These challenges motivate a learning objective that operates in a shared latent space and an alignment mechanism that explicitly accounts for physiological delay.

To address these challenges, we introduce \textbf{\SysName}, a physiology-aware joint-embedding predictive architecture for unified cardiac representation learning across ECG, PPG, and PCG. The model maps heterogeneous waveforms into a common token space with lightweight modality-specific stems and processes them using a single shared Transformer encoder. Since the modalities do not share waveform appearance but do share latent cardiac structure, \SysName learns by predicting masked latent cardiac states rather than reconstructing raw signals. This predictive formulation avoids emphasizing low-level sensor morphology and also removes the need for negative pairs, which can be ambiguous in physiological data because different subjects or segments may share rhythm, heart rate, or pathology.

\SysName is also designed around the data regime of cardiac sensing. Large unimodal ECG, PPG, and PCG corpora are increasingly available, but synchronized multi-sensor recordings are comparatively scarce. We therefore use a two-stage curriculum. The model first learns within-modality structure from abundant unimodal recordings through intra-modal masked latent prediction. It then uses paired recordings to align modalities through delay-aware cross-modal prediction, while continuing to sample unimodal data to preserve per-modality performance. Additional physiology-guided objectives encourage cardiac-cycle phase awareness, consistent delay estimates, and robust shared representations.

We evaluate \SysName as a frozen encoder across 25 downstream tasks spanning ECG, PPG, and PCG, where a single shared encoder consistently outperforms strong modality-specific baselines.
Overall, our contributions are as follows:
\begin{itemize}
    \item We formulate unified cardiac representation learning as the problem of recovering a shared latent cardiac state from heterogeneous ECG, PPG, and PCG observations that are shifted in time by physiological delay.
    \item We propose \SysName, a joint-embedding predictive architecture that learns shared cardiac representations through intra-modal masked latent prediction and delay-aware cross-modal latent prediction.
    \item We introduce a physiology-guided alignment strategy in which inter-modality delay is explicitly estimated and supervised, allowing electrical, hemodynamic, and acoustic signals to be aligned in latent cardiac time rather than by raw timestamp.
    \item We show that our pretrained encoder transfers across ECG, PPG, and PCG tasks, outperforming modality-specific baselines on average for PPG and PCG while remaining competitive on ECG, providing a unified alternative to separate per-sensor representation learning.
\end{itemize}

\section{Related Work}

\paragraph{General Time Series Foundation Models.}
General time series foundation models have become an important direction for learning transferable representations across forecasting, classification, and anomaly detection tasks~\cite{shi2025time, ansari2025chronos}. Most of these models rely on self-supervised learning to exploit large unlabeled time series corpora. Within this area, contrastive methods (TS2Vec~\cite{yue2022ts2vec}, CoST~\cite{woo2022cost}), masked-prediction methods inspired by masked autoencoders~\cite{mae2022,goswami2024moment}, and joint-embedding predictive architectures~\cite{ennadir2024joint} learn transferable representations by predicting latent rather than raw structure. Despite this progress, most general time series foundation models underperform on focused clinical tasks while still treating sensors or channels as independent streams and therefore do not explicitly model the structured correspondences that arise when multiple modalities observe the same underlying physical process.

\paragraph{Cardiac Signal Foundation Models.}
Building on progress in general self-supervised learning, recent work has focused on cardiac signals as a domain where large unlabeled datasets are increasingly accessible. For ECG, early self-supervised models such as ST-MEM~\cite{na2024guiding} have demonstrated strong transfer to arrhythmia, rhythm, and morphology classification using masked prediction objectives on publicly available recordings. A recent line of ECG foundation models, including ECGFounder~\cite{ecgfounder}, ECG-FM~\cite{mckeen2025ecgfm}, HeartLang~\cite{heartlang}, and D-BETA~\cite{dbeta}, further leverages large-scale paired labels or weak clinical reports as a multimodal framework to achieve higher performance, though at the cost of requiring privileged annotation. For PPG, PaPaGei~\cite{papagei2025} and AnyPPG~\cite{nie2025anyppg} apply contrastive pretraining to wearable recordings and show strong performance in heart rate and blood pressure estimation. For PCG, general audio models such as AudioMAE~\cite{huang2022audiomae} and CLAP~\cite{elizalde2023clap} have been adapted to heart sound classification. However, all of these models are designed for a single sensing modality. \SysName differs by training one shared encoder across ECG, PPG, and PCG simultaneously, treating the physiological correspondence between modalities as an additional supervisory signal rather than a complication to be avoided.

\section{Method}

\begin{figure*}[t]
\centering
\includegraphics[width=0.9\textwidth]{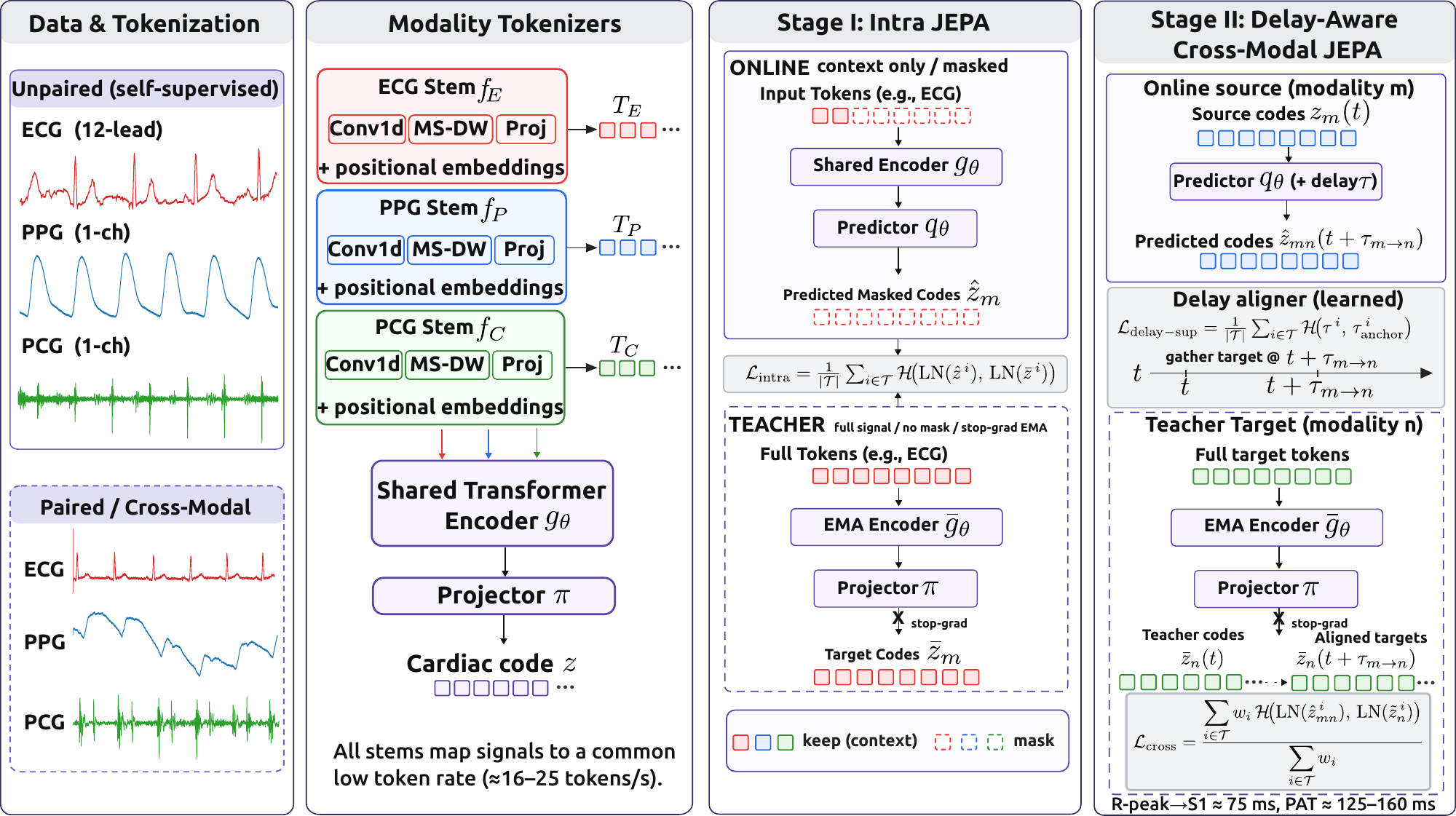}
\caption{Overview of \SysName. Stage~I learns unimodal cardiac codes with masked latent prediction from abundant ECG, PPG, and PCG recordings. Stage~II uses paired recordings for delay-aware cross-modal prediction, where a learned aligner estimates the physiological offset so that one modality predicts another at the corresponding cardiac time.}
\label{fig:architecture}
\end{figure*}

\subsection{Overview}
\SysName learns a single representation that serves ECG, PPG, and PCG together, instead of training a separate model per sensor. As shown in Figure~\ref{fig:architecture}, each modality is first mapped into a common token space by a light modality-specific tokenizer, and all modalities are then processed by one shared Transformer encoder that produces a common cardiac code. Because the modalities are offset in time by physiology, we train in two stages (detailed below): Stage~I learns each modality's structure from abundant unimodal data, and Stage~II aligns modalities from scarce paired data through a learned delay.

\subsection{Problem Formulation}
We view ECG, PPG, and PCG not as three signal families but as three renderings of one hidden process, which makes precise what a shared representation should capture. A cardiac cycle produces a latent state $\mathbf{c}(t)\in\mathbb{R}^{d_c}$, and each sensor observes it through its own transduction and at its own point along the cardiovascular path:
\begin{equation}
    \mathbf{x}_m(t) \;=\; \mathcal{O}_m\!\big(\mathbf{c}\big(t-\tau_m(t)\big),\; \mathbf{u}_m(t)\big),
    \label{eq:generative}
\end{equation}
where $\mathcal{O}_m$ is a modality-specific rendering, $\mathbf{u}_m$ collects sensor nuisance such as noise and placement, and $\tau_m$ is a physiological delay from electrical activation. These delays set cardiac signals apart from ordinary multimodal data: the ECG is nearly immediate, the PCG follows electromechanical coupling, and the PPG arrives only after pulse transit. Here, two modalities of the same beat thus share $\mathbf{c}$ but differ in rendering, nuisance, and relative delay $\tau_{m\rightarrow n}=\tau_n-\tau_m$. Two requirements follow: First, since $\mathbf{c}$ is common to the whole cycle, the shared representation must be recoverable from a partial view of one modality. Second, since different sensors describe the same state at the same instant, their representations must agree once the delay $\tau_{m\rightarrow n}$ is removed. We do not model $\mathbf{u}_m$ explicitly; predicting in latent space rather than reconstructing waveforms, together with pressure toward modality invariance, is enough to reduce reliance on it. The two requirements map onto the two stages: the first is met from abundant unimodal data, the second from the scarce paired data.

\subsection{Modality Tokenizers and Shared Encoder}
Since ECG, PPG, and PCG differ in sampling rate, channels, and morphology, we make them comparable with a modality-specific tokenizer $f_m$ before the shared encoder, so the encoder spends its capacity on cardiac structure rather than acquisition differences. Given $\mathbf{x}_m\in\mathbb{R}^{C_m\times T_m}$, the tokenizer produces $\mathbf{h}_m=f_m(\mathbf{x}_m)\in\mathbb{R}^{N\times d}$: a strided convolution whose stride is set per modality so all three emit tokens at a comparable rate, a multi-scale depthwise block that adds context around short events such as the QRS complex, PPG upstroke, and first heart sound, and a linear projection, followed by sinusoidal positional and modality embeddings. A single shared Transformer encoder then processes these tokens with no modality-specific layers, so ECG, PPG, and PCG update the same parameters. Writing $\mathbf{H}_m=g_\theta(\mathbf{h}_m)$, a shared projector maps each token to the cardiac code $\mathbf{Z}_{m}=\pi(\mathbf{H}_m)$ used for both training and transfer. A momentum encoder $\bar g_{\bar\theta}$, an exponential moving average of the tokenizers, encoder, and projector, is run without gradients on the unmasked signal to supply stable targets.

\subsection{Stage~I: Intra-Modal JEPA}
Stage~I enforces the first requirement, recovering the cardiac code from a partial view of one modality. We predict masked regions in latent space rather than reconstructing the waveform, because reconstruction rewards fitting sensor noise and fine morphology, exactly the nuisance we want the code to forget. Following the joint-embedding predictive principle~\cite{ijepa2023}, we mask a large fraction of the tokens in contiguous blocks, encode the visible context, and let a light predictor $q_\theta$ match the momentum encoder's code at the masked positions,
\begin{equation}
    \mathcal{L}_\text{intra}
    =\frac{1}{|\mathcal{T}|}\sum_{i\in\mathcal{T}}
    \mathcal{H}\!\big(\mathrm{LN}(\hat{\mathbf{z}}^{\,i}),\,\mathrm{LN}(\bar{\mathbf{z}}^{\,i})\big) ,
    \label{eq:intra}
\end{equation}
where $\hat{\mathbf{z}}^{\,i}$ is the prediction, $\bar{\mathbf{z}}^{\,i}$ the stop-gradient target, $\mathrm{LN}$ LayerNorm, and $\mathcal{H}$ the Smooth-L1 loss. Because the masked blocks routinely span whole beats, the model cannot rely on local interpolation and must infer the missing state from neighboring cycles, which drives the code toward the rhythm and phase structure shared across modalities.

\subsection{Stage~II: Delay-Aware Cross-Modal JEPA}
Intra-modal prediction recovers the code within each sensor but does not make sensors agree, the second requirement. Stage~II predicts one modality from another on paired recordings, introduced after Stage~I because alignment is meaningful only once each modality has reliable structure. Aligning tokens at equal timestamps would match an electrical event to a mechanical or hemodynamic one and teach surface appearance rather than shared state, so we predict through a learned delay. A delay head $h_\delta$ maps the source code $\mathbf{z}^{i}_{m}$ and a target-modality embedding $\mathbf{e}_{n}$ to a bounded per-token offset:
\begin{equation}
\label{eq:delay}
\tau^{i}_{m\rightarrow n} = \tau_{\max}\tanh\big(h_\delta([\mathbf{z}^{i}_{m}; \mathbf{e}_{n}])\big).
\end{equation}
and the target is gathered with a Gaussian kernel centered at the shifted time $t_i+\tau^{\,i}$ over the in-band target tokens,
\begin{equation}
    a_{ij}=\mathrm{softmax}_j\!\Big(-\tfrac12\big(\tfrac{t_j-t_i-\tau^{\,i}}{\sigma}\big)^2\Big),
    \qquad
    \tilde{\mathbf{z}}_n^{\,i}=\sum_j a_{ij}\,\bar{\mathbf{z}}_n^{\,j}.
    \label{eq:gather}
\end{equation}
A soft kernel is used rather than nearest-token selection because it is differentiable in $\tau^{\,i}$ and thus gives the delay head a usable gradient, and it reduces to a soft interpolation when $\sigma$ is comparable to the token spacing. Each position is weighted by the certainty of its alignment,
\begin{equation}
    w_i = 1 - H(a_i)/\log N_i \;\;(N_i>1),\qquad w_i=0\;\;(N_i\le1),
    \label{eq:conf}
\end{equation}
with $H$ the entropy of $a_i$ and $N_i$ the number of in-band tokens, so peaked, well-covered alignments approach one and ambiguous or empty ones are downweighted or excluded. The cross-modal objective predicts the aligned target under these weights,
\begin{equation}
    \mathcal{L}_\text{cross}
    =\frac{\sum_{i\in\mathcal{T}} w_i\,
    \mathcal{H}\!\big(\mathrm{LN}(\hat{\mathbf{z}}_{mn}^{\,i}),\,\mathrm{LN}(\tilde{\mathbf{z}}_{n}^{\,i})\big)}
    {\sum_{i\in\mathcal{T}} w_i + \epsilon} ,
    \label{eq:cross}
\end{equation}
where $\hat{\mathbf{z}}_{mn}^{\,i}$ comes from the same predictor $q_\theta$ as in Eq.~\eqref{eq:intra}, now conditioned on $\tau^{\,i}$; sharing the predictor keeps both objectives in one space, with the two differing only in how the predictor is conditioned. Left unconstrained, the delay drifts toward trivial solutions, so we anchor it to physiology: beat detection gives a reference offset per pair, the R-peak to first-heart-sound interval for ECG--PCG and the pulse arrival time for ECG--PPG, and we regress the estimate onto it,
\begin{equation}
    \mathcal{L}_\text{delay-sup}
    =\frac{1}{|\mathcal{T}|}\sum_{i\in\mathcal{T}}\mathcal{H}\!\big(\tau^{\,i},\,\tau^{\,i}_{\text{anchor}}\big) .
    \label{eq:delaysup}
\end{equation}
This supervision is applied only where reliable anchors exist; when too few clean beats are detected, the term is masked out while the delay head still runs and the window still contributes to $\mathcal{L}_\text{cross}$, so noisy detection removes only the label, never the alignment path. The delay is therefore learned and physiologically supervised rather than fixed, and it is the mechanism that encourages a single shared space across electrical, acoustic, and hemodynamic sensing.

\subsection{Auxiliary Objectives and Training}
Our framework leverages two light terms to further shape the code: a cross-modal VICReg~\cite{bardes2022vicreg} term $\mathcal{L}_\text{state}$ pulls paired modalities into the same space while preventing collapse, and a phase term $\mathcal{L}_\text{phase}$ predicts intra-beat phase from anchors, tying the code to the cardiac cycle. The full objective is:
\begin{equation}
\begin{aligned}
    \mathcal{L}
    =
    &\lambda_\text{intra}\mathcal{L}_\text{intra}
    + \lambda_\text{cross}\mathcal{L}_\text{cross}
    + \lambda_\text{delay}\mathcal{L}_\text{delay-sup} \\
    &+ \lambda_\text{state}\mathcal{L}_\text{state}
    + \lambda_\text{phase}\mathcal{L}_\text{phase}.
\end{aligned}
\end{equation}
The curriculum follows the data asymmetry. In Stage~I, each sample is a single modality, so only intra-modal prediction and the phase term are active; the cross-modal, delay, and state terms require paired inputs and switch on in Stage~II, which warm-starts from Stage~I and spends the scarce paired data on alignment while keeping unimodal data in the mixture so per-modality codes do not drift. At test time the encoder is frozen: a signal passes through its tokenizer, the shared encoder, and the projector, and the pooled code is read by a linear head, so every result measures transfer without encoder adaptation.

\begin{table*}[t]
\centering
{\footnotesize
\setlength{\tabcolsep}{10pt}
\begin{tabular}{lcccccc}
\toprule
\textbf{Task} & \textbf{PaPaGei-S} & \textbf{PaPaGei-P} & \textbf{AnyPPG} & \textbf{PulsePPG} & \textbf{ChronosBolt} & \textbf{\SysName} \\
\midrule
\multicolumn{7}{l}{\textit{PPG Classification Tasks}} \\[2pt]
WESAD $\uparrow$            & 65.3 & 65.7 & \underline{70.1} & 66.8 & 64.0 & \textbf{74.5} \\
DaLiA Activity $\uparrow$   & 68.4 & 71.1 & \underline{79.0} & 77.3 & 77.2 & \textbf{90.3} \\
MIMIC AF $\uparrow$         & 48.4 & 81.1 & \underline{93.3} & 32.4 & 46.1 & \textbf{97.7} \\
PPG Arrhythmia $\uparrow$   & 86.3 & 88.7 & \underline{95.8} & 92.6 & 90.4 & \textbf{96.8} \\
BIDMC RR $\uparrow$         & 51.3 & 38.2 & \underline{53.2} & 43.5 & 38.0 & \textbf{78.4} \\
UQVital RR $\uparrow$       & 40.6 & \underline{41.8} & \underline{41.8} & 29.5 & 30.5 & \textbf{44.6} \\
\cmidrule(lr){1-7}
\textit{Average (Classification)} $\uparrow$ & 60.1 & 64.4 & \underline{72.2} & 57.0 & 57.7 & \textbf{80.4} \\
\midrule
\multicolumn{7}{l}{\textit{PPG Regression Tasks}} \\[2pt]
DaLiA HR $\downarrow$       & 11.8 & 11.5 & \underline{5.8}  & 8.6  & 8.2  & \textbf{3.8} \\
EarSet HR $\downarrow$      & 10.7 & 9.5  & 9.8  & 8.9  & \underline{8.3}  & \textbf{8.2} \\
WildPPG HR $\downarrow$     & 10.0 & 9.9  & \textbf{5.4} & 7.7  & \underline{7.5}  & \textbf{5.4} \\
Sensors DBP $\downarrow$    & 7.2  & 7.0  & \underline{6.9} & 7.4  & \underline{6.9} & \textbf{6.8} \\
Sensors SBP $\downarrow$    & 20.0 & 19.4 & \textbf{16.7} & 18.6 & 18.0 & \underline{16.8} \\
UCI DBP $\downarrow$        & 7.8  & 7.9  & \underline{6.7}  & 6.9  & 7.2  & \textbf{5.8} \\
UCI SBP $\downarrow$        & 18.5 & 18.1 & \underline{15.2} & 16.2 & 17.4 & \textbf{10.3} \\
BCG DBP $\downarrow$        & 10.1 & 12.9 & \underline{4.5}  & 11.2 & 10.8 & \textbf{2.8} \\
BCG SBP $\downarrow$        & 13.7 & \underline{10.2} & 13.5 & 12.7 & 11.8 & \textbf{4.4} \\
UQVital SpO$_2$ $\downarrow$ & \textbf{0.76} & 0.85 & \underline{0.78} & 0.84 & 0.83 & 0.82 \\
WildPPG RMSSD $\downarrow$  & 36.6 & 37.6 & \underline{34.9} & 36.7 & 35.9 & \textbf{34.8} \\
\cmidrule(lr){1-7}
\textit{Average (Regression)} $\downarrow$ & 13.4 & 13.2 & \underline{10.9} & 12.3 & 12.1 & \textbf{9.1} \\
\bottomrule
\end{tabular}}
\caption{PPG Baseline Linear Probe Results (100\% training data). $\uparrow$: macro-AUROC $\times 100$ (higher is better); $\downarrow$: MAE (lower is better). \textbf{Bold}: best; \underline{underline}: second best. Average (Classification) is the unweighted mean macro-AUROC across the 6 classification tasks. Average (Regression) is the unweighted mean MAE across the 11 regression tasks.}
\label{tab:ppg_baselines}
\end{table*}

\section{Experiments}

\subsection{Datasets and Downstream Tasks}
In the first stage, the unimodal pretraining dataset includes MIMIC-IV-ECG~\cite{PhysioNet-mimic-iv-ecg-1.0} with 800K 12-lead ECG recordings at 500\,Hz, PPG-EXT~\cite{mimicppg2026} as a large-scale PPG corpus at 125\,Hz, and BMD-HS~\cite{ali2026bmdhs} for PCG recordings at 4000\,Hz. In the second stage, we introduce cross-modal supervision by further using PPG-EXT~\cite{mimicppg2026} and VitalDB~\cite{vitaldb2022} for synchronous ECG-PPG pairs, EPHNOGRAM~\cite{ephnogram2024} for ECG-PCG pairs, and SensSmartTech~\cite{senssmarttech2024} for trimodal recordings. Full dataset statistics are reported in the Appendix.

After pretraining, we assess whether the learned representation transfers to various downstream tasks. For ECG, we evaluate on PTB-XL~\cite{wagner2020ptb} (with four subgroups: superclass, subclass, form, and rhythm) for arrhythmia classification, together with CPSC 2018~\cite{liu2018open} and CSN~\cite{zheng202012}, following~\cite{liu2024zero}. For PPG, we use seventeen tasks covering arrhythmia detection, stress and activity recognition, respiratory rate, heart rate regression, blood pressure regression, SpO$_2$ estimation, and HRV estimation, using the pre-processed data from PulseLM~\cite{pham2026pulselm}. For PCG, we evaluate murmur detection on CirCor DigiScope~\cite{PhysioNet-circor-heart-sound-1.0.3} and abnormal heart sound detection on CinC2016~\cite{cinc2016}. The signal samples can be found in the Appendix. Finally, classification tasks are reported using macro-AUROC $\times 100$, while regression tasks are reported using MAE.

\subsection{Baselines and Evaluation Protocol}
Given the broad range of downstream tasks, we compare \SysName against baselines specific to each modality. For ECG, we include ten self-supervised ECG-only methods: SimCLR~\cite{chen2020simple}, BYOL~\cite{grill2020bootstrap}, BarlowTwins~\cite{zbontar2021barlow}, MoCo-v3~\cite{chen2021empirical}, SimSiam~\cite{chen2021exploring}, TS-TCC~\cite{eldele2021time}, CLOCS~\cite{kiyasseh2021clocs}, ASTCL~\cite{wang2023adversarial}, CRT~\cite{zhang2023self}, and ST-MEM~\cite{na2024guiding}, following~\cite{liu2024zero}. These models form the main comparison group because, like \SysName, they do not rely on clinical reports or labels during representation learning. Meanwhile, we also report ECGFounder~\cite{ecgfounder} and ECG-Text models, including ECG-FM~\cite{mckeen2025ecgfm}, HeartLang~\cite{heartlang}, AnyChat~\cite{anychat}, ESI~\cite{esi}, MERL~\cite{liu2024zero}, and D-BETA~\cite{dbeta}. These models are useful references, but they are not directly comparable because they use either large-scale supervised labels or paired clinical reports. Therefore, in Table~\ref{tab:ecg_baselines}, bold and underline are assigned only within the self-supervised ECG-only group, with \SysName included in that group. The supervised and text-supervised models are shown separately to indicate how close a signal-only model can come to models trained with additional privileged information.

For PPG, we compare against PaPaGei-S, PaPaGei-P~\cite{papagei2025}, AnyPPG~\cite{nie2025anyppg}, PulsePPG~\cite{saha2025pulseppg}, and ChronosBolt~\cite{chronos2024}. For PCG, we use CLAP~\cite{elizalde2023clap}, AudioMAE~\cite{huang2022audiomae} and StethoLM~\cite{wang2026stetholm} as the main baselines. All methods are evaluated under a linear probing protocol where the pretrained encoder is frozen and only a task-specific linear head is trained. All downstream evaluations are performed using patient-disjoint splits. To further test label efficiency, ECG tasks are evaluated with 1\%, 10\%, and 100\% of the training labels.

\begin{table*}[t]
\centering
\resizebox{\textwidth}{!}{
\begin{tabular}{l ccc ccc ccc ccc ccc ccc c}
\toprule
& \multicolumn{3}{c}{PTB-XL Super} & \multicolumn{3}{c}{PTB-XL Sub} & \multicolumn{3}{c}{PTB-XL Form} & \multicolumn{3}{c}{PTB-XL Rhythm} & \multicolumn{3}{c}{CPSC} & \multicolumn{3}{c}{CSN} & \\
\cmidrule(lr){2-4}\cmidrule(lr){5-7}\cmidrule(lr){8-10}\cmidrule(lr){11-13}\cmidrule(lr){14-16}\cmidrule(lr){17-19}
Method & 1\% & 10\% & 100\% & 1\% & 10\% & 100\% & 1\% & 10\% & 100\% & 1\% & 10\% & 100\% & 1\% & 10\% & 100\% & 1\% & 10\% & 100\% & Avg \\
\midrule
\multicolumn{20}{l}{\textit{Self-supervised ECG-only models}} \\[2pt]
SimCLR         & 63.4 & 69.8 & 73.5 & 60.8 & 68.3 & 73.4 & 55.0 & 57.0 & 62.5 & 51.4 & 69.4 & 77.7 & 59.8 & 68.5 & 76.5 & 59.0 & 67.3 & 73.2 & 65.9 \\
BYOL           & 71.7 & 73.8 & 76.5 & 57.2 & 67.4 & 71.6 & 48.7 & 61.6 & 70.8 & 42.0 & \underline{74.4} & 77.2 & 60.9 & 74.4 & 78.8 & 54.2 & 71.9 & 74.7 & 67.1 \\
BarlowTwins    & 72.9 & 76.0 & 78.4 & \underline{62.6} & 70.8 & 74.3 & 52.1 & 60.4 & 66.1 & 50.1 & 73.5 & 77.6 & 55.1 & 72.8 & 78.4 & \underline{60.7} & 71.6 & 77.4 & 68.4 \\
MoCo-v3        & \underline{73.2} & 76.7 & 78.3 & 55.9 & 69.2 & 76.7 & 50.3 & \underline{63.7} & 71.3 & 51.4 & 71.7 & 74.3 & \underline{62.1} & 76.7 & 75.3 & 54.6 & \underline{74.3} & 77.7 & \underline{68.5} \\
SimSiam        & \underline{73.2} & 72.7 & 75.6 & 62.5 & 69.3 & 76.4 & 55.2 & 62.9 & 71.3 & 49.3 & 69.5 & 75.9 & 58.4 & 72.9 & 75.3 & 58.3 & 68.6 & 77.4 & 68.0 \\
TS-TCC         & 70.7 & 75.9 & 78.9 & 53.5 & 67.0 & 77.9 & 48.0 & 61.8 & 71.2 & 43.3 & 69.5 & \underline{78.2} & 57.1 & 73.6 & 78.7 & 55.3 & 68.5 & 76.8 & 67.0 \\
CLOCS          & 68.9 & 73.4 & 76.3 & 57.9 & \underline{72.6} & 76.2 & 52.0 & 58.0 & \underline{72.7} & 47.2 & 71.9 & 76.3 & 59.6 & \underline{77.8} & 77.5 & 54.4 & 71.9 & 76.1 & 67.8 \\
ASTCL          & 72.5 & 77.3 & \underline{81.0} & 61.9 & 68.8 & 76.5 & 44.1 & 60.9 & 67.0 & \underline{52.4} & 72.0 & 76.1 & 57.9 & 77.0 & 79.5 & 56.4 & 70.9 & 75.8 & 68.2 \\
CRT            & 69.7 & \underline{78.2} & 77.2 & 62.0 & 70.8 & \underline{78.7} & 46.4 & 59.5 & 68.7 & 47.4 & 73.5 & 74.4 & 58.0 & 76.4 & \underline{82.0} & 56.2 & 73.7 & \underline{78.8} & 68.4 \\
ST-MEM         & 61.1 & 66.9 & 71.4 & 54.1 & 57.9 & 63.6 & \underline{55.7} & 60.0 & 66.1 & 51.1 & 65.4 & 74.9 & 56.7 & 63.3 & 70.4 & 59.8 & 66.9 & 71.4 & 63.2 \\
\midrule
\multicolumn{20}{l}{\textit{Label-supervised ECG model}} \\[2pt]
ECGFounder     & 83.1 & 87.2 & 89.6 & 73.6 & 76.5 & 81.7 & 61.1 & 69.7 & 85.1 & 87.0 & 91.6 & 94.6 & 87.6 & 93.6 & 96.3 & 71.9 & 81.7 & 91.9 & 83.5 \\
\midrule
\multicolumn{20}{l}{\textit{ECG-Text foundation models}} \\[2pt]
MERL           & 82.4 & 86.3 & 88.7 & 64.9 & 80.6 & 84.7 & 58.3 & 72.4 & 79.7 & 53.3 & 82.9 & 88.3 & 70.3 & 85.3 & 90.6 & 66.6 & 82.7 & 88.0 & 78.1 \\
HeartLang      & 78.9 & 84.4 & 86.7 & 69.8 & 75.1 & 83.9 & 56.4 & 66.1 & 79.5 & 72.6 & 77.3 & 91.1 & 71.1 & 84.8 & 91.3 & 68.4 & 76.1 & 89.8 & 78.0 \\
AnyChat        & 82.1 & 84.4 & 86.8 & 72.1 & 73.2 & 82.2 & 58.4 & 69.7 & 77.1 & 73.1 & 84.7 & 94.5 & 83.7 & 88.9 & 92.6 & 71.2 & 79.5 & 89.9 & 80.2 \\
ESI            & 72.3 & 80.8 & 83.7 & 61.6 & 69.0 & 75.6 & 56.9 & 59.3 & 72.8 & 56.6 & 71.4 & 81.7 & 68.2 & 78.9 & 83.8 & 60.1 & 65.1 & 79.4 & 71.0 \\
D-BETA         & 83.2 & 88.4 & 90.1 & 77.7 & 82.9 & 85.2 & 70.1 & 78.9 & 84.0 & 86.6 & 92.8 & 96.7 & 85.5 & 91.4 & 94.9 & 80.0 & 87.4 & 90.7 & 85.9 \\
ECG-FM         & 81.3 & 87.0 & 89.4 & 71.7 & 76.9 & 83.0 & 64.6 & 76.3 & 86.1 & 77.0 & 91.0 & 96.3 & 84.7 & 92.6 & 95.7 & 76.0 & 87.7 & 95.0 & 84.0 \\
\midrule
\SysName (Ours)    & \textbf{81.3} & \textbf{86.4} & \textbf{89.1} & \textbf{71.0} & \textbf{79.1} & \textbf{86.4} & \textbf{60.5} & \textbf{70.0} & \textbf{83.9} & \textbf{87.8} & \textbf{92.4} & \textbf{96.4} & \textbf{87.6} & \textbf{93.4} & \textbf{95.8} & \textbf{75.1} & \textbf{83.1} & \textbf{93.8} & \textbf{84.1} \\
\bottomrule
\end{tabular}
}
\caption{ECG linear probing results under 1\%, 10\%, and 100\% label fractions across six datasets. \textbf{Avg} is the mean macro-AUROC across all 18 settings (six datasets $\times$ three label fractions). Label-supervised and ECG-text models are shown as reference models but are excluded from the SSL bold/underline comparison.}
\label{tab:ecg_baselines}
\end{table*}

\subsection{Implementation Details}
Our model is implemented in PyTorch and uses a ViT-B shared encoder (12 layers, 12 attention heads, hidden dimension $d=768$), trained on a single NVIDIA H100 GPU with AdamW. Stage~I runs for 300K steps and Stage~II for 200K steps with the cross-modal objectives enabled. Full optimization hyperparameters (learning-rate schedules, batch sizes, weight decay, and EMA momentum) are provided in the Appendix.

\subsection{Main Results}
We first examine whether a single frozen encoder can remain competitive across all three modalities. Tables~\ref{tab:ppg_baselines}, \ref{tab:ecg_baselines}, and \ref{tab:pcg_baselines} summarize the PPG, ECG, and PCG results, respectively. Overall, \SysName performs consistently across the three sensing domains, despite using the same shared encoder for all downstream tasks.

On PPG tasks, \SysName improves the average classification AUROC from 72.2 to 80.4 compared with the strongest baseline, while also reducing the average regression MAE from 10.9 to 9.1. The gains are especially clear on activity recognition, atrial fibrillation detection, respiratory rate estimation, and blood pressure estimation. For ECG, \SysName improves the average AUROC across all 18 settings by 15.5 points over the strongest self-supervised baseline (MoCo-v3; 84.1 vs.\ 68.5 in Table~\ref{tab:ecg_baselines}), outperforming every self-supervised ECG-only model on nearly all datasets and label fractions. While text-supervised and label-supervised ECG models still provide strong reference points, \SysName matches or surpasses them on several benchmarks without using clinical reports or large supervised ECG labels. For PCG, the model achieves the best results on both CirCor murmur detection and CinC2016 abnormal heart-sound detection under the linear probing protocol. Since the same encoder is also used for ECG and PPG, this result supports the central hypothesis that shared cardiac pretraining can benefit acoustic heart-sound representation, rather than only electrical or optical signals.

\begin{table}[t]
\centering
{\footnotesize
\setlength{\tabcolsep}{9pt}
\begin{tabular}{lcc}
\toprule
\textbf{Method} & \textbf{CirCor} $\uparrow$ & \textbf{CinC2016} $\uparrow$ \\
\midrule
StethoLM &	64.9 $\pm$ 1.2	& 45.7 $\pm$ 0.4 \\
CLAP         & 75.3 $\pm$ 1.8           & 44.3 $\pm$ 0.7          \\
AudioMAE     & \underline{79.1} $\pm$ 2.7 & \underline{62.4} $\pm$ 0.7 \\
\midrule
\SysName (Ours)  & \textbf{97.9} $\pm$ 0.1  & \textbf{66.8} $\pm$ 0.4 \\
\bottomrule
\end{tabular}}
\caption{PCG linear probe results on the murmur detection and abnormal heart sound detection tasks (macro-AUROC $\times 100$, $\uparrow$; mean$\pm$std over 3 seeds).}
\label{tab:pcg_baselines}
\end{table}

\begin{table*}[t]
\centering
{\footnotesize
\setlength{\tabcolsep}{16pt}
\begin{tabular}{lcccc}
\toprule
\textbf{Model / Variant} & \textbf{ECG Avg} $\uparrow$ & \textbf{PPG Cls Avg} $\uparrow$ & \textbf{PPG Reg Avg} $\downarrow$ & \textbf{PCG Avg} $\uparrow$ \\
\midrule
\multicolumn{5}{l}{\textit{Pretraining Modality Combination}} \\[2pt]
ECG only                    & 89.7 $\pm$ 0.5 & --   & --   & --   \\
PPG only                    & --   & 78.5 $\pm$ 2.6 & 10.7 $\pm$ 0.6 & --   \\
PCG only                    & --   & --   & --   & 60.2 $\pm$ 0.9 \\
ECG + PPG                   & \textbf{91.4} $\pm$ 0.4 & \textbf{80.6} $\pm$ 2.8 & 10.1 $\pm$ 0.8 & --   \\
ECG + PCG                   & 88.4 $\pm$ 0.6 & --   & --   & 79.7 $\pm$ 0.7 \\
\cmidrule(lr){1-5}
\textbf{\SysName (Ours)} & 90.9 $\pm$ 0.3 & 80.4 $\pm$ 2.5 & \textbf{9.1} $\pm$ 0.6 & \textbf{82.3} $\pm$ 0.2 \\
\midrule
\multicolumn{5}{l}{\textit{Self-Supervised Objective}} \\[2pt]
SimCLR                  & 89.5 $\pm$ 0.5 & 70.1 $\pm$ 2.9 & 11.1 $\pm$ 0.6 & 80.0 $\pm$ 0.7 \\
BYOL                    & 89.3 $\pm$ 0.4 & 66.7 $\pm$ 3.8 & 11.0 $\pm$ 0.5 & 80.3 $\pm$ 0.4 \\
BarlowTwins             & 88.6 $\pm$ 0.4 & 64.1 $\pm$ 4.4 & 11.3 $\pm$ 0.5 & 78.8 $\pm$ 1.0 \\
MAE                     & 84.4 $\pm$ 0.5 & 62.0 $\pm$ 3.8 & 11.8 $\pm$ 0.5 & 78.1 $\pm$ 0.3 \\
\cmidrule(lr){1-5}
\textbf{\SysName (Ours)} & \textbf{90.9} $\pm$ 0.3 & \textbf{80.4} $\pm$ 2.5 & \textbf{9.1} $\pm$ 0.6 & \textbf{82.3} $\pm$ 0.2 \\
\midrule
\multicolumn{5}{l}{\textit{Auxiliary Loss Terms}} \\[2pt]
w/o Cross-modal prediction          & 83.8 $\pm$ 0.5 & 63.9 $\pm$ 3.5 & 10.7 $\pm$ 0.5 & 78.4 $\pm$ 0.4 \\
w/o State alignment                 & 84.4 $\pm$ 0.5 & 66.3 $\pm$ 3.5 & 11.2 $\pm$ 0.5 & 79.3 $\pm$ 0.5 \\
w/o Delay modeling                  & 85.3 $\pm$ 0.6 & 68.9 $\pm$ 3.7 & 11.1 $\pm$ 0.6 & 81.3 $\pm$ 0.5 \\
w/o Phase supervision               & 88.7 $\pm$ 0.5 & 71.0 $\pm$ 3.1 & 10.4 $\pm$ 0.6 & 78.3 $\pm$ 0.5 \\
\cmidrule(lr){1-5}
\textbf{\SysName (Ours)}         & \textbf{90.9} $\pm$ 0.3 & \textbf{80.4} $\pm$ 2.5 & \textbf{9.1} $\pm$ 0.6 & \textbf{82.3} $\pm$ 0.2 \\
\bottomrule
\end{tabular}}
\caption{Ablation studies on pretraining modalities, self-supervised objectives, and auxiliary losses.}
\label{tab:ablation}
\end{table*}

\subsection{Visualizing the Shared Space}
\begin{figure}[t]
\centering
\includegraphics[width=0.9\columnwidth]{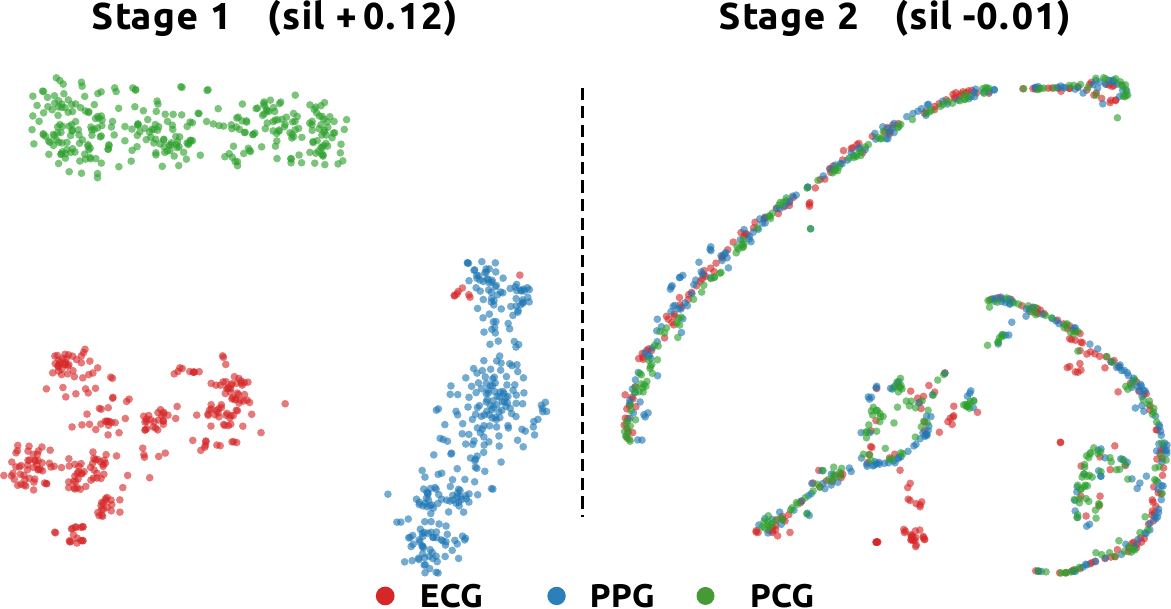}
\caption{T-SNE of cardiac codes from co-recorded ECG, PPG, and PCG samples after Stage~I (left) and Stage~II (right).}
\label{fig:tsne}
\end{figure}

While the downstream results show task-level transfer, they do not directly reveal how the shared cardiac space is organized. To inspect this, we visualize the pooled cardiac codes of held-out co-recorded ECG, PPG, and PCG samples using t-SNE at two points in training, after Stage~I and after Stage~II. Figure~\ref{fig:tsne} shows that after Stage~I the samples form clear modality-specific clusters, since each modality has only been trained in isolation. After Stage~II the same samples become thoroughly interleaved across modalities, indicating that cross-modal training reshapes the space from a sensor-driven layout into a modality-invariant one.

We quantify this shift with the modality silhouette of the pooled codes, which measures how separable the ECG, PPG, and PCG clusters are. It falls from $0.121$ after Stage~I to $-0.006$ after Stage~II, so the sensor-specific separation is effectively removed, and the shared code becomes largely invariant to the sensing modality. Because downstream performance also improves from Stage~I to Stage~II, this mixing reflects genuine cross-modal alignment rather than a collapse of the representation.

\subsection{Ablation Studies}
Finally, we analyze which parts of \SysName contribute to the observed gains. We start with the role of multimodal pretraining. The top block of Table~\ref{tab:ablation} compares models pretrained with different modality combinations while keeping the architecture and training budget fixed. This comparison allows us to separate the effect of additional cardiac modalities from simple changes in model capacity. The full trimodal setting gives the best overall balance across the three domains, strongest on PPG regression and PCG and within noise of the best bimodal variant on ECG and PPG classification, indicating that the added modalities provide useful supervisory signal for single-modality downstream performance rather than a simple capacity increase.

To further isolate the effect of the learning objective, we replace the JEPA objective with SimCLR~\cite{chen2020simple}, BYOL~\cite{grill2020bootstrap}, BarlowTwins~\cite{zbontar2021barlow}, and MAE~\cite{mae2022}, while keeping the data and architecture unchanged. As shown in the middle block of Table~\ref{tab:ablation}, masked latent prediction gives the strongest overall results. This supports the use of JEPA for heterogeneous cardiac signals, where predicting latent structure is likely more useful than reconstructing raw waveforms or relying only on instance discrimination.

We also remove each auxiliary objective from the full model in the bottom block of Table~\ref{tab:ablation}. The results show that no single term explains all of the improvements. Instead, cross-modal prediction, state alignment, delay modeling, and phase supervision provide complementary benefits. This is consistent with the design of \SysName, where the main prediction losses learn transferable representations while the auxiliary losses guide the representation toward physiologically meaningful structure. The Appendix further reports an input-length ablation and a sensitivity analysis over the loss weights, both of which show that \SysName is robust to these choices.

\section{Conclusion}
We studied whether cardiac representation learning can move beyond separate sensor-specific pretraining toward a shared foundation model of latent cardiac physiology. We introduced \SysName, a physiology-aware joint embedding predictive framework that learns from ECG, PPG, and PCG through intra-modal latent prediction and delay-aware cross-modal alignment. By predicting latent cardiac states and aligning modalities at the corresponding cardiac time, \SysName enables electrical, hemodynamic, and acoustic signals to supervise one another. Our experiments show that the unified model transfers across modalities and surpasses strong self-supervised signal baselines on representative tasks. These results support our hypothesis that heterogeneous cardiac signals can provide mutual supervision for learning a shared physiological representation. We hope this motivates a shift from per-sensor cardiac models toward unified representations that exploit, rather than avoid, the physiological correspondence between modalities.

\section*{Acknowledgments}
\small{This work was supported by the NWO AiNed Fellowship Grant of A.S., and in part by Google.org and the Google Cloud Research Credits program through the Gemini Academic Program. We also acknowledge the use of the Dutch National Supercomputer Snellius for essential computational tasks.}

\bibliography{main}
\clearpage
\appendix

\section{Appendix}

\subsection{Data and Training Configuration}
Table~\ref{tab:training_config} summarizes the pretraining and downstream linear probing settings used throughout our experiments. For each pretraining corpus we list the number of training and validation recordings, and for each downstream benchmark we list the number of classes together with the train, validation, and test splits. Every downstream task uses the same optimizer, epoch budget, batch size, and learning rate, so that the reported differences reflect the quality of the frozen encoder rather than any per task tuning of the linear head. Keeping this protocol identical across ECG, PPG, and PCG also makes the comparison between the three sensing modalities easy to interpret, since only the input data and the linear head change from one task to the next. Tasks with a dash in the class column are regression tasks that we evaluate with mean absolute error, while all remaining tasks are classification tasks that we evaluate with macro-AUROC.

\subsection{Additional Implementation Details}
The main paper reports the architecture and the high level training budget, and here we give the full optimization settings so that the results can be reproduced. Both pretraining stages use AdamW with weight decay $0.05$. In Stage~I the model is pretrained for 300K steps with batch size 196 and a peak learning rate of $1.2 \times 10^{-4}$ that cosine decays to $1 \times 10^{-6}$. In Stage~II the model is trained for a further 200K steps with batch size 64 and a peak learning rate of $6 \times 10^{-5}$ that decays to $1 \times 10^{-6}$, with the cross modal objectives enabled. The shared encoder follows the ViT-B configuration with 12 layers, 12 attention heads, and hidden dimension 768. The momentum encoder is an exponential moving average whose momentum increases linearly from $0.998$ to $0.9999$ over the course of pretraining. All pretraining runs on a single NVIDIA H100 GPU. During downstream evaluation the encoder is frozen and only a linear head is trained, using the per task splits and the AdamW settings listed in Table~\ref{tab:training_config}.

\subsection{Training Procedure}
Algorithm~\ref{alg:training} summarizes the two stage curriculum. Stage~I learns the structure of each modality from abundant unimodal data through masked latent prediction, and Stage~II warm starts from Stage~I and spends the scarce paired data on delay aware cross modal prediction while continuing to sample unimodal data so that the per modality codes do not drift.

\begin{table*}[t]
\centering
\setlength{\tabcolsep}{12pt}
\resizebox{\textwidth}{!}{%
\begin{tabular}{lcccccccc}
\toprule
\textbf{Dataset} &
\textbf{\# Classes} &
\textbf{Train} &
\textbf{Valid} &
\textbf{Test} &
\textbf{Opt.} &
\textbf{Epochs} &
\textbf{BS} &
\textbf{LR} \\
\midrule
\multicolumn{9}{l}{\textit{ECG Pretraining}} \\
MIMIC-IV-ECG~\cite{PhysioNet-mimic-iv-ecg-1.0}
& -- & 710,560 & 78,951 & -- & AdamW & -- & 196 & 1.2e-4 \\
\midrule
\multicolumn{9}{l}{\textit{PPG Pretraining}} \\
PPG-EXT~\cite{mimicppg2026}
& -- & 4,611,607 & 512,401 & -- & AdamW & -- & 196 & 1.2e-4 \\
\midrule
\multicolumn{9}{l}{\textit{PCG Pretraining}} \\
BMD-HS~\cite{ali2026bmdhs}
& -- & 3,436 & 382 & -- & AdamW & -- & 196 & 1.2e-4 \\
\midrule
\multicolumn{9}{l}{\textit{ECG Downstream}} \\
PTB-XL Super~\cite{wagner2020ptb}
& 5 & 17,084 & 2,146 & 2,158 & AdamW & 100 & 16 & 1e-3 \\
PTB-XL Sub
& 23 & 17,084 & 2,146 & 2,158 & AdamW & 100 & 16 & 1e-3 \\
PTB-XL Form
& 19 & 7,197 & 901 & 880 & AdamW & 100 & 16 & 1e-3 \\
PTB-XL Rhythm
& 12 & 16,832 & 2,100 & 2,098 & AdamW & 100 & 16 & 1e-3 \\
CPSC 2018~\cite{liu2018open}
& 9 & 4,950 & 551 & 1,376 & AdamW & 100 & 16 & 1e-3 \\
CSN~\cite{zheng202012}
& 38 & 16,546 & 1,860 & 4,620 & AdamW & 100 & 16 & 1e-3 \\
\midrule
\multicolumn{9}{l}{\textit{PPG Downstream}} \\
WESAD
& 2 & 2,104 & 297 & 597 & AdamW & 100 & 16 & 1e-3 \\
DaLiA
& 8 & 29,294 & 4,602 & 5,320 & AdamW & 100 & 16 & 1e-3 \\
MIMIC AF
& 2 & 3,239 & 240 & 717 & AdamW & 100 & 16 & 1e-3 \\
PPG Arrhythmia
& 2 & 36,820 & 4,764 & 5,243 & AdamW & 100 & 16 & 1e-3 \\
BIDMC
& -- & 9,412 & 1,652 & 1,398 & AdamW & 100 & 16 & 1e-3 \\
UQVital
& -- & 22,158 & 2,529 & 8,158 & AdamW & 100 & 16 & 1e-3 \\
EarSet
& -- & 1,368 & 44 & 364 & AdamW & 100 & 16 & 1e-3 \\
WildPPG
& -- & 179,492 & 15,000 & 45,000 & AdamW & 100 & 16 & 1e-3 \\
Sensors BP
& -- & 1,631 & 180 & 250 & AdamW & 100 & 16 & 1e-3 \\
UCI BP
& -- & 89,054 & 11,286 & 11,411 & AdamW & 100 & 16 & 1e-3 \\
BCG BP
& -- & 521 & 86 & 64 & AdamW & 100 & 16 & 1e-3 \\
\midrule
\multicolumn{9}{l}{\textit{PCG Downstream}} \\
CirCor Murmur~\cite{PhysioNet-circor-heart-sound-1.0.3}
& 3 & 1,745 & 608 & 654 & AdamW & 100 & 16 & 1e-3 \\
CinC~\cite{cinc2016}
& 2 & 2,006 & 627 & 607 & AdamW & 100 & 16 & 1e-3 \\
\bottomrule
\end{tabular}%
}
\caption{Training configurations for pretraining and downstream linear probing. Regression tasks are shown with a dash in the class column and are scored with mean absolute error, while classification tasks are scored with macro-AUROC.}
\label{tab:training_config}
\end{table*}

\begin{algorithm}[!ht]
\caption{Two-stage training of \SysName}
\label{alg:training}
\begin{algorithmic}[1]
\REQUIRE Unimodal corpora $\{\mathcal{D}_m\}$ for ECG, PPG, PCG; paired corpora $\mathcal{P}$; tokenizers $\{f_m\}$; shared encoder $g_\theta$; projector $\pi$; predictor $q_\theta$; delay head $h_\delta$; momentum encoder $\bar g_{\bar\theta}$
\STATE \textbf{Stage I, intra-modal prediction}
\FOR{each pretraining step}
  \STATE Sample a unimodal batch $x_m$ from $\{\mathcal{D}_m\}$
  \STATE Tokenize $h_m = f_m(x_m)$ and split into context and masked blocks
  \STATE Encode the context with $g_\theta$ and predict the masked codes with $q_\theta$
  \STATE Form stop-gradient targets with $\bar g_{\bar\theta}$ on the unmasked signal
  \STATE Update $\theta$ on $\lambda_\text{intra}\mathcal{L}_\text{intra} + \lambda_\text{phase}\mathcal{L}_\text{phase}$
  \STATE Update $\bar\theta$ as an EMA of $\theta$
\ENDFOR
\STATE \textbf{Stage II, delay-aware cross-modal prediction}, warm started from Stage~I
\FOR{each pretraining step}
  \STATE With probability $p$ sample a paired batch $(x_m, x_n)$ from $\mathcal{P}$, otherwise a unimodal batch
  \IF{the batch is paired}
    \STATE Estimate the per token delay $\tau_{m\to n} = \tau_{\max}\tanh(h_\delta(z_m))$
    \STATE Gather the aligned target with a Gaussian kernel centered at $t + \tau$
    \STATE Update $\theta$ and $\delta$ on $\lambda_\text{cross}\mathcal{L}_\text{cross} + \lambda_\text{delay}\mathcal{L}_\text{delay-sup} + \lambda_\text{state}\mathcal{L}_\text{state}$
  \ELSE
    \STATE Update $\theta$ on $\lambda_\text{intra}\mathcal{L}_\text{intra} + \lambda_\text{phase}\mathcal{L}_\text{phase}$
  \ENDIF
  \STATE Update $\bar\theta$ as an EMA of $\theta$
\ENDFOR
\ENSURE Frozen shared encoder $g_\theta$ for downstream linear probing
\end{algorithmic}
\end{algorithm}

\subsection{Input Length Ablation}
We further study whether \SysName stays stable when the length of the input window changes, because a representation that is sensitive to the exact segment duration is harder to deploy in practice. Table~\ref{tab:ablation_length} reports four representative ECG task groups at window lengths of 2.5, 5, and 10 seconds, using 10\% of the training labels. For each task group we give the score at every window length together with the mean and the standard deviation across the three settings. \SysName reaches the smallest standard deviation in most task groups, which shows that its cardiac representation stays reliable regardless of how much signal context is available at test time. This robustness matters in real deployments, where the amount of clean signal that a sensor can capture varies with the recording conditions and with the patient.

\begin{table*}[t]
\centering
\resizebox{\textwidth}{!}{
\begin{tabular}{l ccccc ccccc ccccc ccccc}
\toprule
& \multicolumn{5}{c}{\textit{PTB-XL Super $\uparrow$}}
& \multicolumn{5}{c}{\textit{PTB-XL Sub $\uparrow$}}
& \multicolumn{5}{c}{\textit{PTB-XL Form $\uparrow$}}
& \multicolumn{5}{c}{\textit{PTB-XL Rhythm $\uparrow$}} \\
\cmidrule(lr){2-6}\cmidrule(lr){7-11}\cmidrule(lr){12-16}\cmidrule(lr){17-21}
\textbf{Method}
& \textbf{2.5s} & \textbf{5s} & \textbf{10s} & \textit{Mean} & \textit{Std}
& \textbf{2.5s} & \textbf{5s} & \textbf{10s} & \textit{Mean} & \textit{Std}
& \textbf{2.5s} & \textbf{5s} & \textbf{10s} & \textit{Mean} & \textit{Std}
& \textbf{2.5s} & \textbf{5s} & \textbf{10s} & \textit{Mean} & \textit{Std} \\
\midrule
ST-MEM      & 65.3 & 69.5 & 69.2 & \textit{68.0} & \textit{1.9} & 58.9 & 60.7 & 60.0 & \textit{59.9} & \textit{0.7} & 52.4 & 58.1 & 53.5 & \textit{54.7} & \textit{2.5} & 62.6 & 58.7 &
62.2 & \textit{61.2} & \textit{\textbf{1.8}} \\
ECG-FM      & 80.5 & 81.9 & 86.5 & \textit{83.0} & \textit{2.6} & 73.3 & 72.2 & 75.1 & \textit{73.5} & \textit{1.2} & 62.2 & 67.8 & \textbf{76.0} & \textit{\textbf{68.7}} & \textit{5.7}
& 72.3 & 77.0 & 82.0 & \textit{77.1} & \textit{4.0} \\
ECGFounder  & 83.3 & \textbf{86.5} & \textbf{87.0} & \textit{85.6} & \textit{1.6} & 74.8 & 76.0 & 76.2 & \textit{75.7} & \textit{0.6} & 62.6 & 68.2 & 69.3 & \textit{66.7} & \textit{2.9}
& 73.5 & 81.7 & 90.2 & \textit{81.8} & \textit{6.8} \\
\midrule
\SysName & \textbf{85.2} & 85.6 & 86.4 & \textit{\textbf{85.7}} & \textit{\textbf{0.5}} & \textbf{78.4} & \textbf{78.7} & \textbf{79.1} & \textit{\textbf{78.7}} &
\textit{\textbf{0.3}} & \textbf{65.6} & \textbf{68.3} & 70.0 & \textit{68.0} & \textit{\textbf{1.8}} & \textbf{87.0} & \textbf{87.3} & \textbf{92.4} & \textit{\textbf{88.9}} &
\textit{2.5} \\
\bottomrule
\end{tabular}
}
\caption{Input length ablation on ECG tasks (macro-AUROC $\times 100$, $\uparrow$, 10\% training data). Bold marks the best value in each column. Mean and Std are the mean and standard deviation across the three input lengths.}
\label{tab:ablation_length}
\end{table*}

\subsection{Sensitivity to Loss Weights}
The full objective combines the intra-modal and cross-modal prediction losses with three auxiliary terms whose relative weights are $\lambda_\text{cross}$, $\lambda_\text{delay}$, and $\lambda_\text{state}$. To check that our results do not depend on a narrow choice of these weights, we vary each one around the value used by \SysName while holding the architecture, the data, and all other weights fixed, and we re-evaluate the frozen encoder. Tables~\ref{tab:sens_cross}, \ref{tab:sens_delay}, and \ref{tab:sens_state} report the four aggregate metrics for the cross modal weight, the delay supervision weight, and the state weight respectively, with the \SysName configuration ($\lambda_\text{cross}=1$, $\lambda_\text{delay}=1$, $\lambda_\text{state}=0.05$) shown as the reference in each table. ECG Avg is the mean macro-AUROC over the six ECG datasets at full labels, PPG Cls is the mean over the six PPG classification tasks, PPG Reg is the mean MAE over the eleven regression tasks, and PCG Avg is the mean over the two PCG tasks.

\begin{table*}[t]
\centering
{\footnotesize
\setlength{\tabcolsep}{5pt}
\begin{tabular}{lcccc}
\toprule
\textbf{Variant} & \textbf{ECG Avg} $\uparrow$ & \textbf{PPG Cls} $\uparrow$ & \textbf{PPG Reg} $\downarrow$ & \textbf{PCG Avg} $\uparrow$ \\
\midrule
\textbf{\SysName} ($\lambda_\text{cross}=1.0$) & \textbf{90.9} & \textbf{80.4} & \textbf{9.1} & \textbf{82.3} \\
$\lambda_\text{cross}=0.5$            & 90.9 & 78.5 & 10.34 & 79.4 \\
$\lambda_\text{cross}=2.0$            & 90.6 & 77.3 & 10.54 & 80.9 \\
\bottomrule
\end{tabular}}
\caption{Sensitivity to the cross-modal loss weight $\lambda_\text{cross}$. The \SysName row uses the default weighting. Values other than MAE are macro-AUROC $\times 100$, where higher is better, and lower is better for MAE.}
\label{tab:sens_cross}
\end{table*}

\begin{table*}[t]
\centering
{\footnotesize
\setlength{\tabcolsep}{5pt}
\begin{tabular}{lcccc}
\toprule
\textbf{Variant} & \textbf{ECG Avg} $\uparrow$ & \textbf{PPG Cls} $\uparrow$ & \textbf{PPG Reg} $\downarrow$ & \textbf{PCG Avg} $\uparrow$ \\
\midrule
\textbf{\SysName} ($\lambda_\text{delay}=1.0$) & \textbf{90.9} & \textbf{80.4} & \textbf{9.1} & 82.3 \\
$\lambda_\text{delay}=0.5$            & 90.8 & 78.7 & 9.68 & 79.3 \\
$\lambda_\text{delay}=2.0$            & 90.8 & 78.0 & 10.20 & 82.5 \\
\bottomrule
\end{tabular}}
\caption{Sensitivity to the delay supervision loss weight $\lambda_\text{delay}$. The \SysName row uses the default weighting. Values other than MAE are macro-AUROC $\times 100$, where higher is better, and lower is better for MAE.}
\label{tab:sens_delay}
\end{table*}

\begin{table*}[t]
\centering
{\footnotesize
\setlength{\tabcolsep}{5pt}
\begin{tabular}{lcccc}
\toprule
\textbf{Variant} & \textbf{ECG Avg} $\uparrow$ & \textbf{PPG Cls} $\uparrow$ & \textbf{PPG Reg} $\downarrow$ & \textbf{PCG Avg} $\uparrow$ \\
\midrule
\textbf{\SysName} ($\lambda_\text{state}=0.05$) & \textbf{90.9} & \textbf{80.4} & \textbf{9.1} & 82.3 \\
$\lambda_\text{state}=0.25$            & 90.7 & 74.5 & 10.20 & 82.7 \\
$\lambda_\text{state}=1.0$             & 90.2 & 75.8 & 9.65 & 82.7 \\
\bottomrule
\end{tabular}}
\caption{Sensitivity to the state loss weight $\lambda_\text{state}$. The \SysName row uses the default weighting. Values other than MAE are macro-AUROC $\times 100$, where higher is better, and lower is better for MAE.}
\label{tab:sens_state}
\end{table*}

Across all three tables the default configuration used by \SysName gives the best overall balance. It attains the strongest PPG classification score of $80.4$ and the lowest PPG regression error of $9.1$, and it matches the best ECG average of $90.9$, while remaining within a few tenths of a point of the best PCG result. No perturbation improves more than one metric at a time, and the settings that raise PCG by a small margin, namely a larger state weight, do so at a clear cost to PPG classification. The ECG average is nearly constant across every setting, changing by less than one point as each weight is varied by a factor of two or more, which shows that the electrical representation is essentially insensitive to the exact loss weighting. PPG classification and regression are the most responsive metrics, and both are best at the default, which indicates that the chosen weights are well matched to the wearable tasks. PCG improves slightly as the state and delay weights increase, which is consistent with the role of these terms in pulling the acoustic modality toward the shared cardiac state, although the gain is small and does not outweigh the loss elsewhere. Taken together, these results show that \SysName is robust to the exact loss weighting and that the default configuration is the best single choice across the three sensing modalities.

\raggedbottom
\subsection{Representative Signal Examples}
\label{app:signal_examples}
We group the representative waveforms by the role they play in \SysName and present each group in its own figure, so that the reader can compare the modalities within a single training stage before comparing across stages.

\begin{figure}[!tbp]
\centering
\scriptsize
\renewcommand{\arraystretch}{1.1}
\begin{tabular}{c}
\includegraphics[width=0.72\columnwidth,keepaspectratio,trim={0mm 0mm 0mm 6mm},clip]{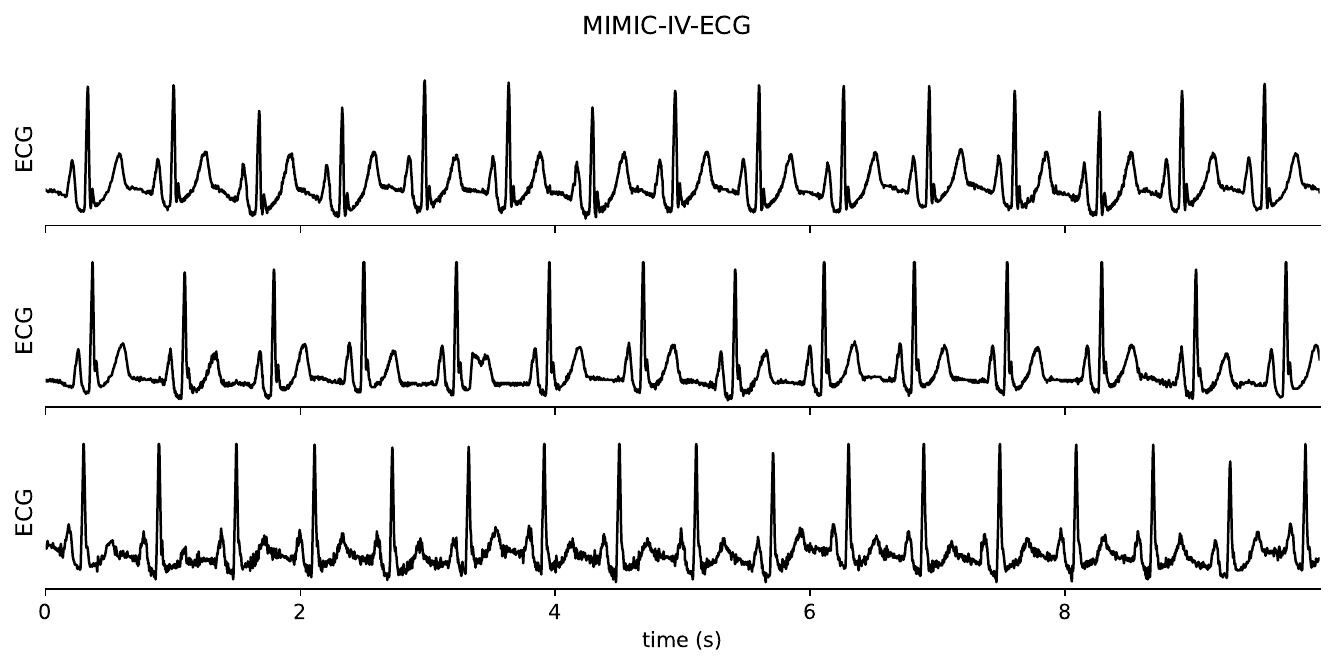} \\[-1mm]
(a) ECG \\[2mm]
\includegraphics[width=0.72\columnwidth,keepaspectratio,trim={0mm 0mm 0mm 6mm},clip]{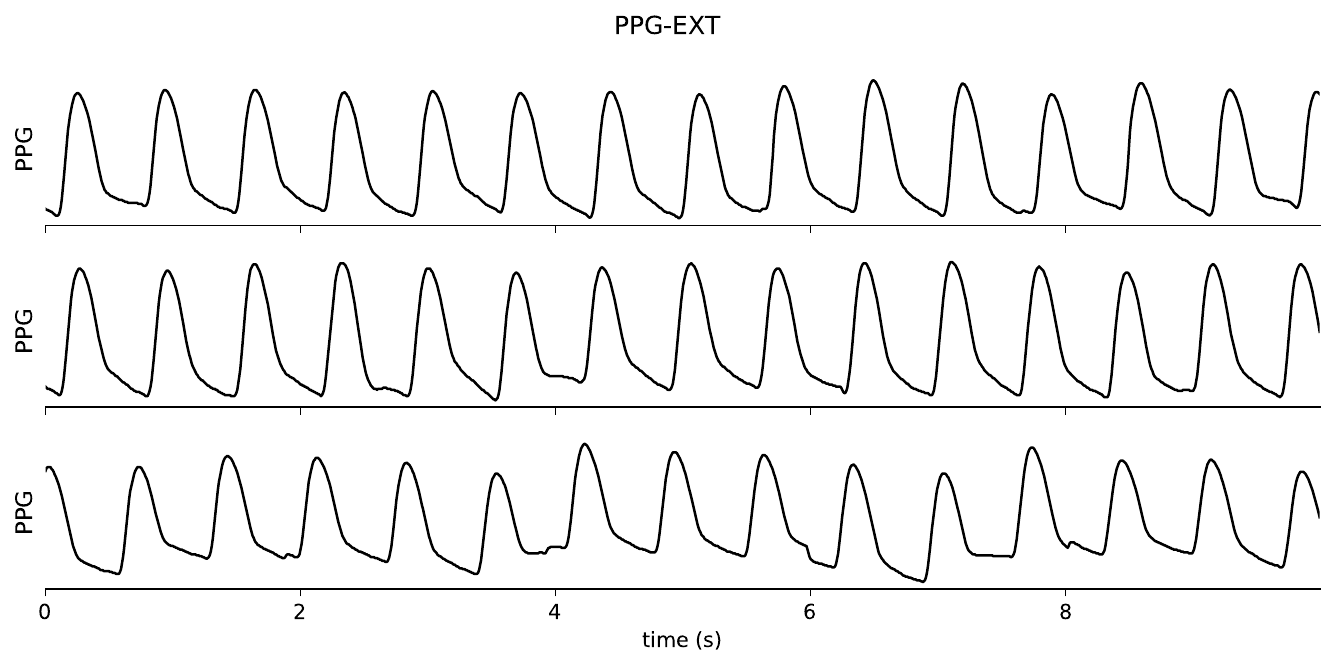} \\[-1mm]
(b) PPG \\[2mm]
\includegraphics[width=0.72\columnwidth,keepaspectratio,trim={0mm 0mm 0mm 6mm},clip]{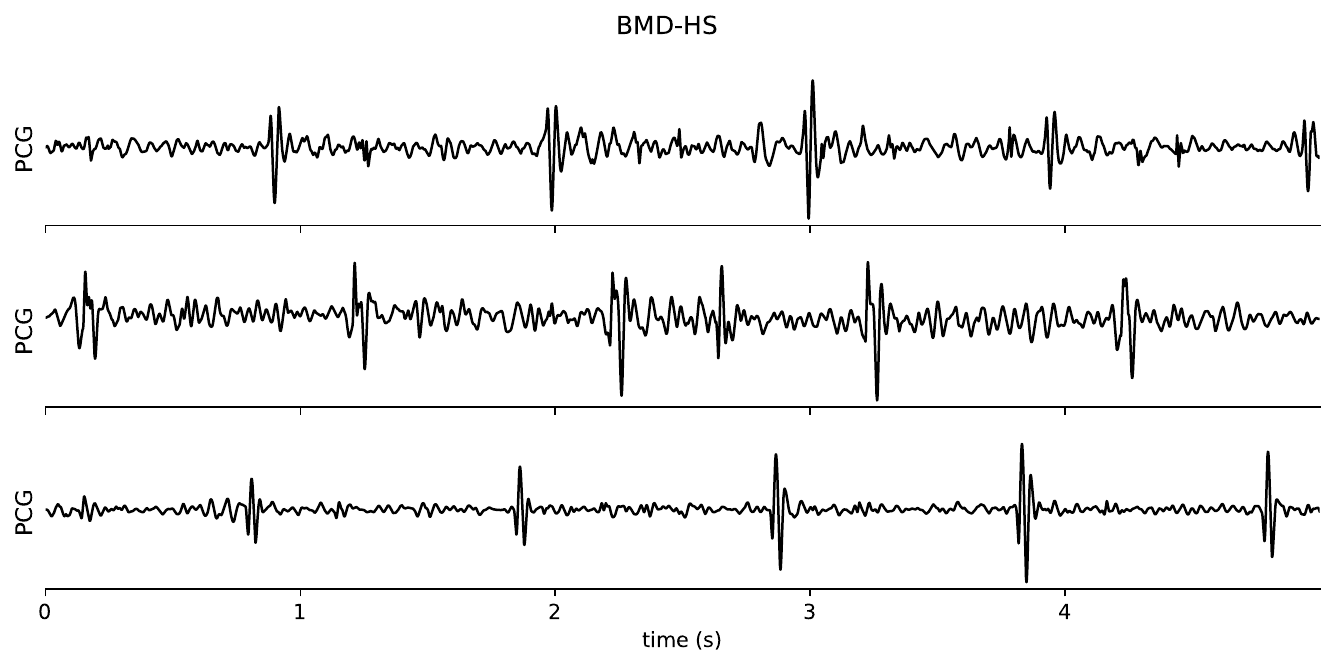} \\[-1mm]
(c) PCG \\
\end{tabular}
\caption{Unimodal pretraining signals used in Stage~I, shown one modality per row. ECG, PPG, and PCG recordings observe the same cardiac cycle through electrical, hemodynamic, and acoustic sensing, yet differ in morphology, sampling rate, and noise pattern.}
\label{fig:sig_pretrain}
\end{figure}

Figure~\ref{fig:sig_pretrain} shows the unimodal pretraining examples for ECG, PPG, and PCG. These recordings supply the abundant single modality data used in Stage~I, where the model learns the structure of each modality on its own through masked latent prediction. Even within this group the three sensors look nothing alike, since each one renders the cardiac cycle through a different physical process, yet all of them carry the same rhythm and timing information that the shared encoder is trained to recover.

\begin{figure}[!tbp]
\centering
\scriptsize
\renewcommand{\arraystretch}{1.1}
\begin{tabular}{c}
\includegraphics[width=0.72\columnwidth,keepaspectratio,trim={0mm 0mm 0mm 6mm},clip]{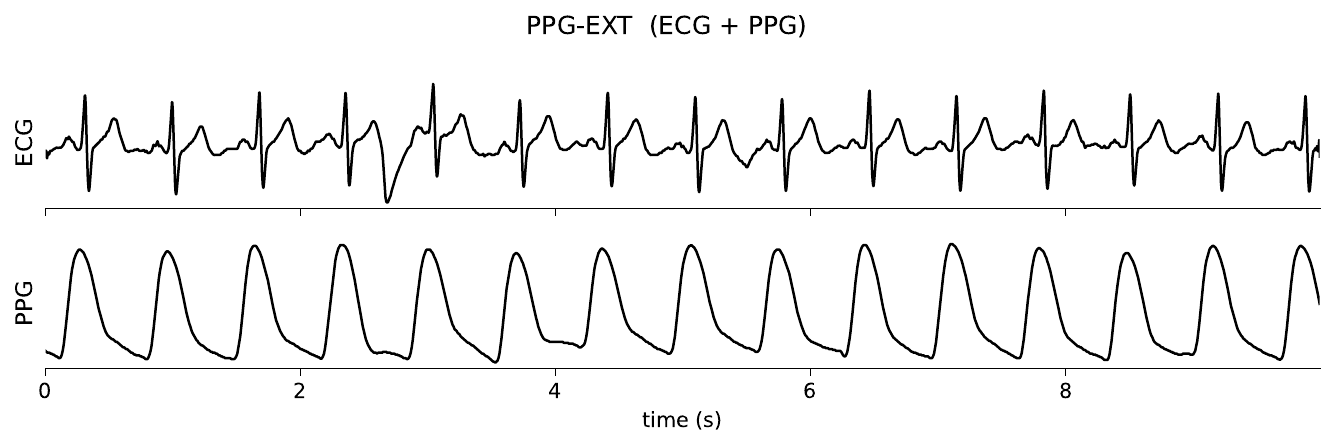} \\[-1mm]
(a) ECG and PPG \\[2mm]
\includegraphics[width=0.72\columnwidth,keepaspectratio,trim={0mm 0mm 0mm 6mm},clip]{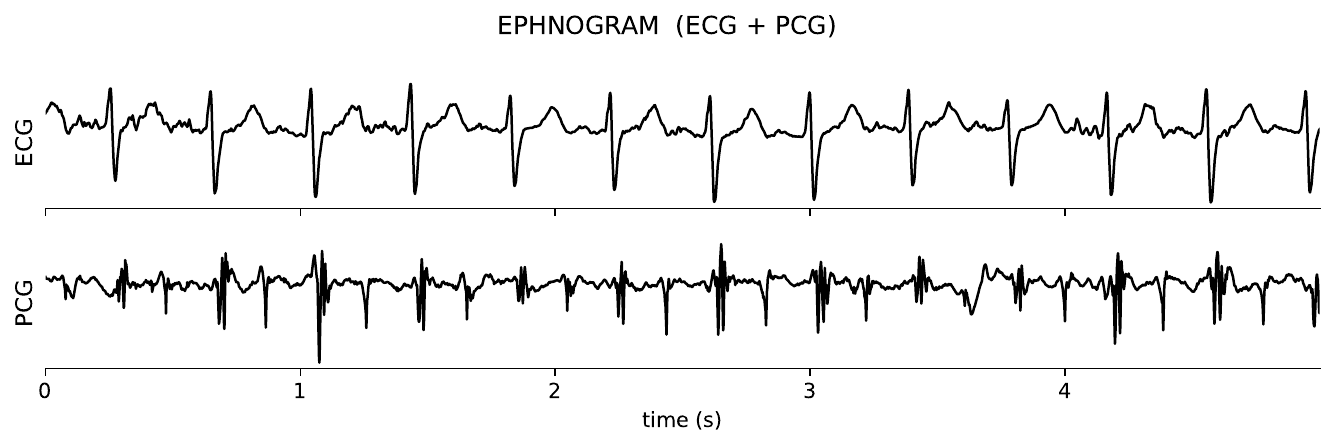} \\[-1mm]
(b) ECG and PCG \\[2mm]
\includegraphics[width=0.72\columnwidth,keepaspectratio,trim={0mm 0mm 0mm 6mm},clip]{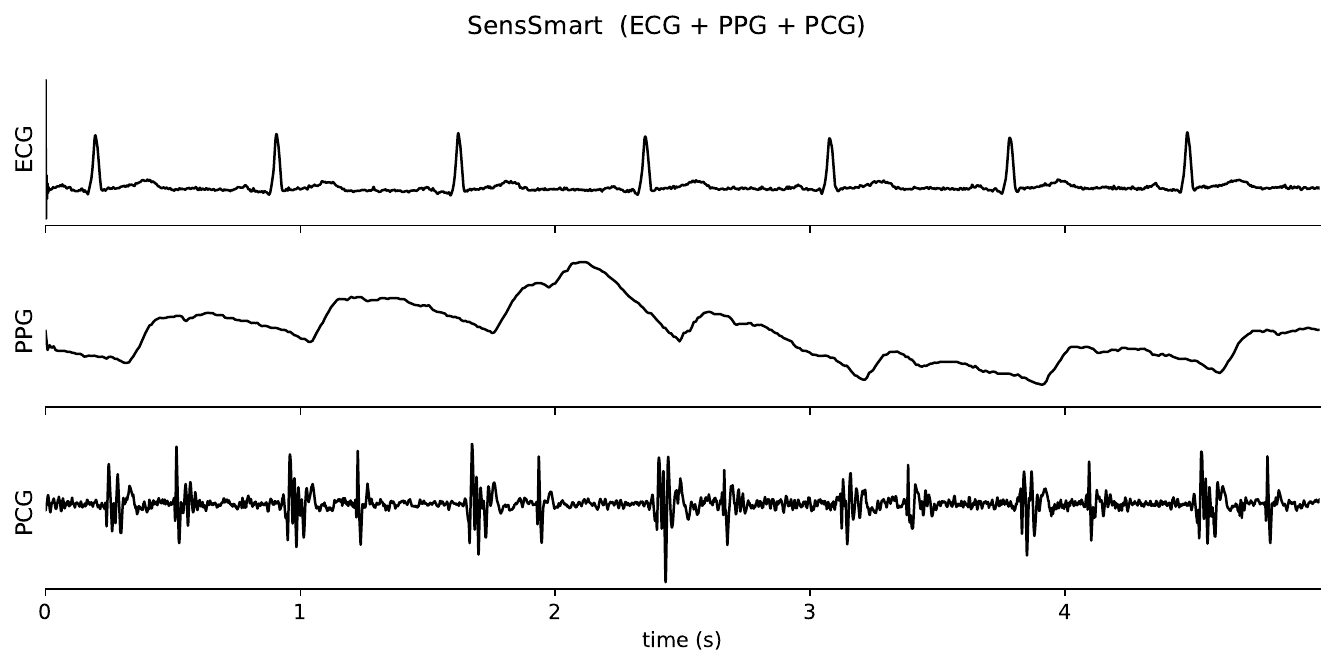} \\[-1mm]
(c) Trimodal \\
\end{tabular}
\caption{Synchronized paired and trimodal recordings used for Stage~II delay aware cross modal alignment, shown one pairing per row. The same beat appears at a different time in each sensor, which motivates the learned physiological delay.}
\label{fig:sig_cross}
\end{figure}

Figure~\ref{fig:sig_cross} shows the synchronized paired and trimodal recordings used in Stage~II. These co-recorded segments are far scarcer than the unimodal data, and they are the only source that ties the modalities together in time. The figure makes clear why alignment by raw timestamp is insufficient, since the same beat appears at a visibly different position in each sensor, which is exactly the offset that the learned delay aligner is designed to correct.

\begin{figure}[!tbp]
\centering
\scriptsize
\renewcommand{\arraystretch}{1.1}
\begin{tabular}{c}
\includegraphics[width=0.72\columnwidth,keepaspectratio,trim={0mm 0mm 0mm 6mm},clip]{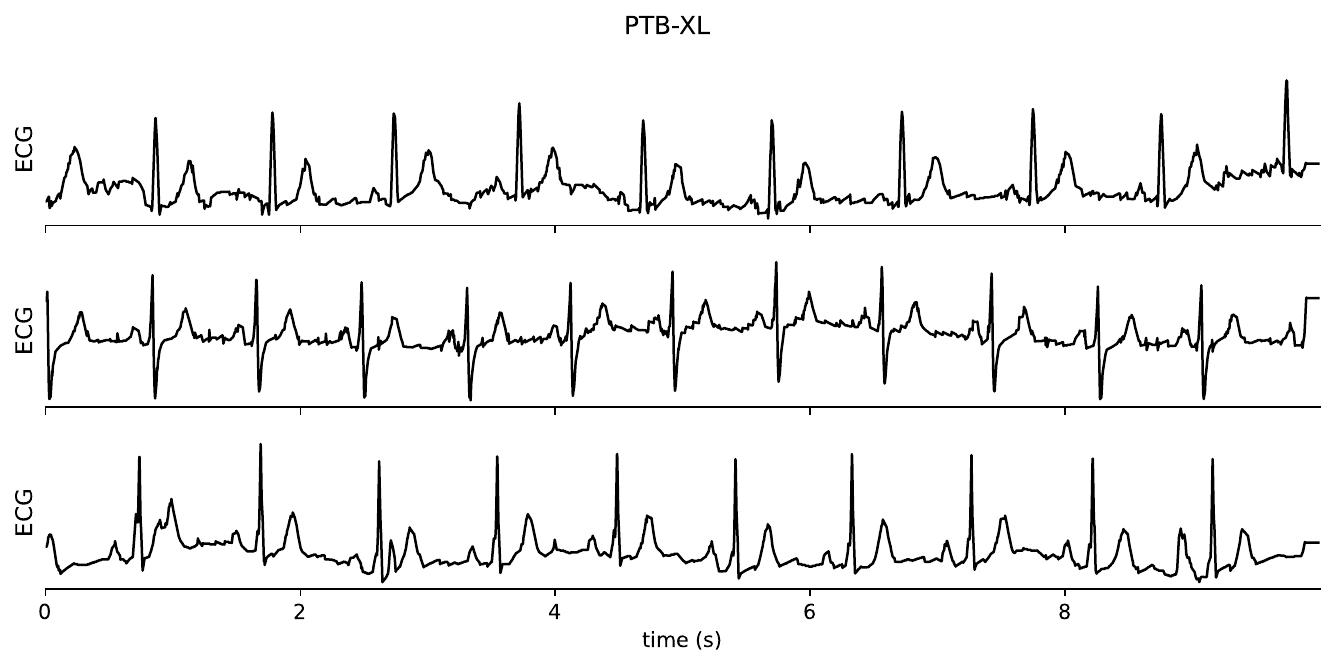} \\[-1mm]
(a) ECG \\[2mm]
\includegraphics[width=0.72\columnwidth,keepaspectratio,trim={0mm 0mm 0mm 6mm},clip]{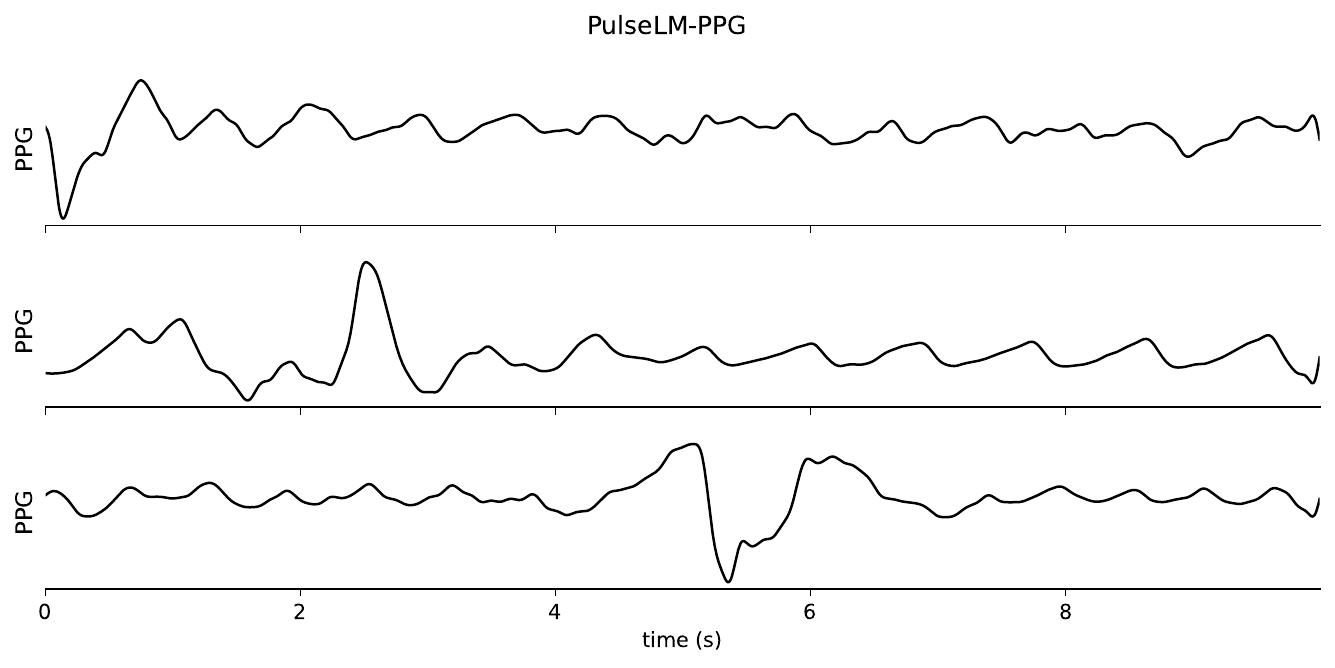} \\[-1mm]
(b) PPG \\[2mm]
\includegraphics[width=0.72\columnwidth,keepaspectratio,trim={0mm 0mm 0mm 6mm},clip]{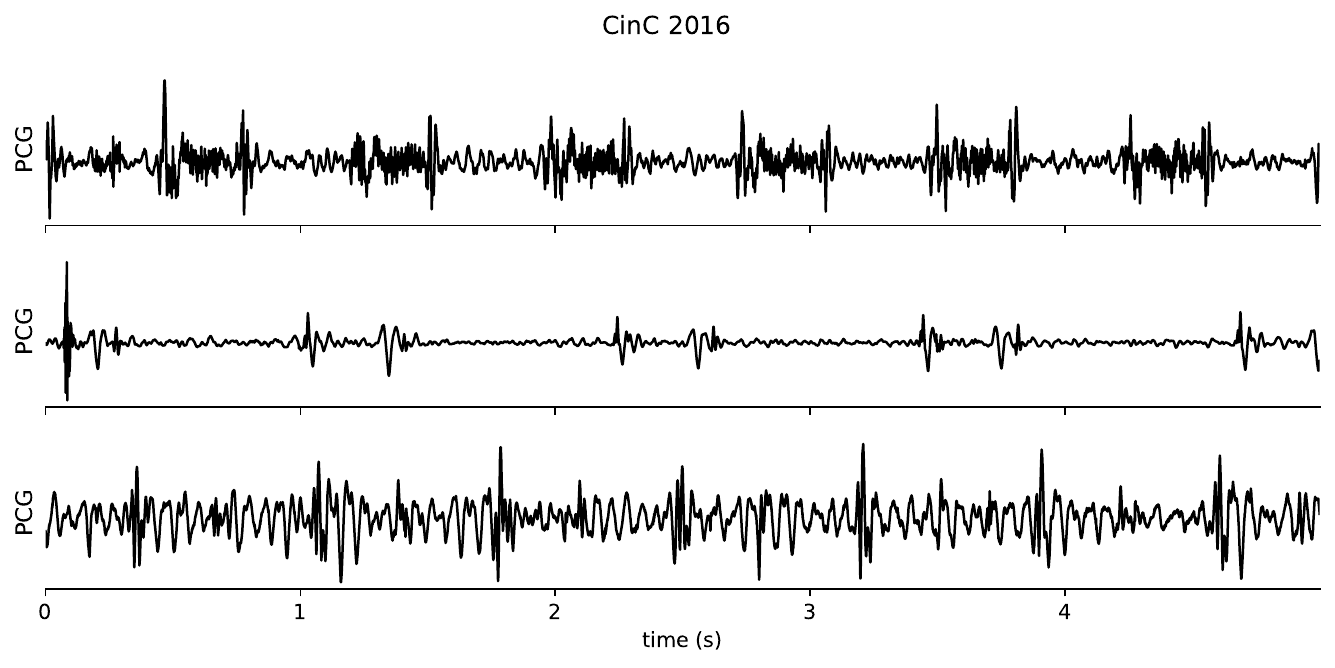} \\[-1mm]
(c) PCG \\
\end{tabular}
\caption{Downstream evaluation signals, shown one modality per row. These held out ECG, PPG, and PCG segments probe the frozen encoder and are never seen during pretraining.}
\label{fig:sig_down}
\end{figure}

Figure~\ref{fig:sig_down} shows the downstream evaluation examples drawn from the benchmarks listed in Table~\ref{tab:training_config}. These segments are never seen during pretraining and are used only to probe the frozen encoder. Although they differ from the pretraining data in morphology, sampling rate, and noise pattern, they describe the same underlying cardiac process, which is the property that lets a single shared representation transfer across all three sensing modalities.

\subsection{Attention Maps on Cardiac Signals}
\label{app:attention_maps}
\begin{figure}[!tbp]
\centering
\scriptsize
\renewcommand{\arraystretch}{1.2}
\setlength{\tabcolsep}{0pt}
\begin{tabular}{c}
\includegraphics[width=0.98\columnwidth,keepaspectratio,trim={0mm 0mm 0mm 0mm},clip]{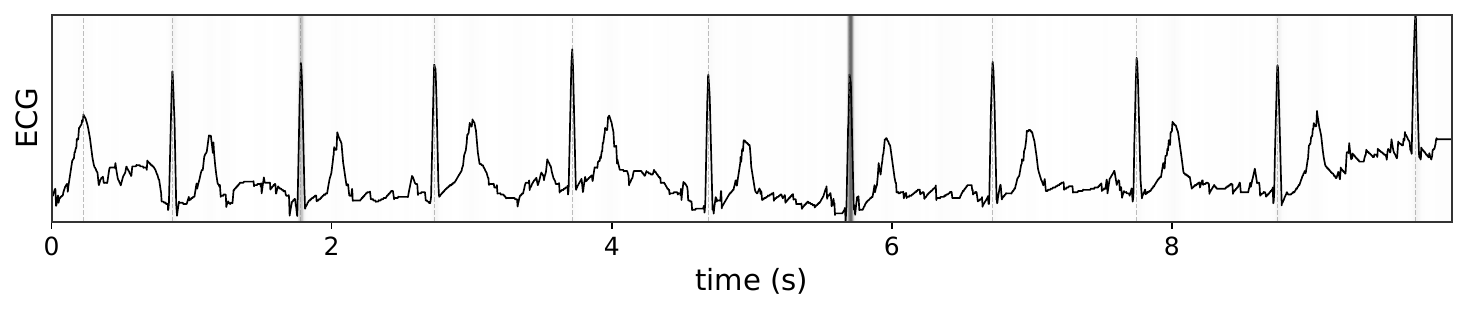} \\[-1mm]
(a) ECG attention map \\[2mm]
\includegraphics[width=0.98\columnwidth,keepaspectratio,trim={0mm 0mm 0mm 0mm},clip]{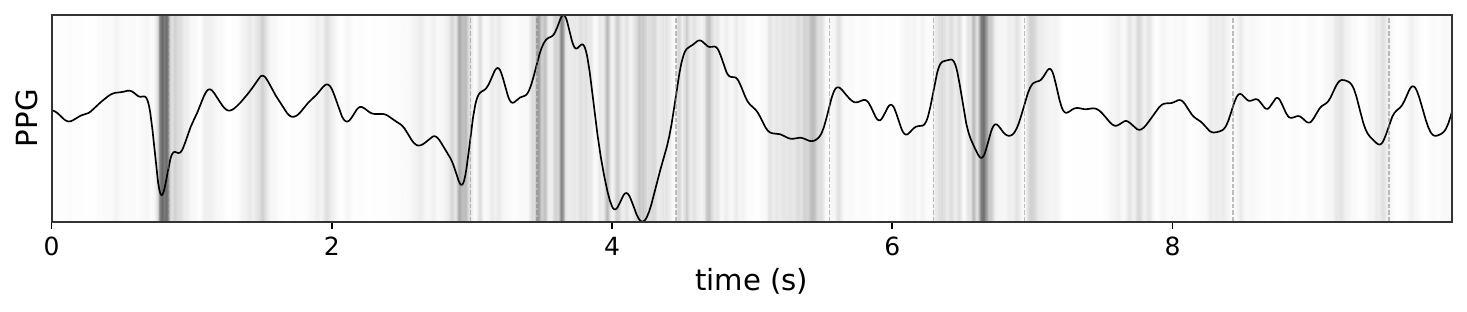} \\[-1mm]
(b) PPG attention map \\[2mm]
\includegraphics[width=0.98\columnwidth,keepaspectratio,trim={0mm 0mm 0mm 0mm},clip]{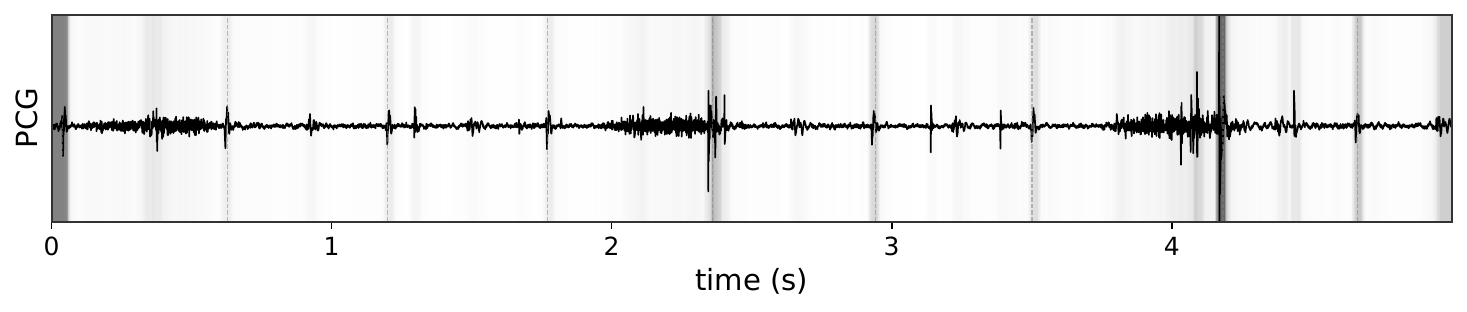} \\[-1mm]
(c) PCG attention map \\
\end{tabular}
\caption{Attention visualizations from \SysName on representative ECG, PPG, and PCG signals. The ECG map highlights sharp electrical events such as QRS complexes, the PPG map emphasizes pulse morphology including the upstroke and systolic peak, and the PCG map focuses on localized acoustic heart-sound events. These patterns indicate that the shared encoder attends to modality specific manifestations of the cardiac cycle while learning a common cardiac representation.}
\label{fig:attention_maps_appendix}
\end{figure}

Figure~\ref{fig:attention_maps_appendix} visualizes attention maps from \SysName on representative ECG, PPG, and PCG signals. The examples show that the shared encoder concentrates on physiologically meaningful regions in every modality, namely sharp depolarization complexes in ECG, the pulse upstroke and systolic peak in PPG, and localized acoustic bursts in PCG. Because a single set of parameters produces all three maps, the figure also shows that the encoder has learned to recognize the modality specific signature of the same cardiac events rather than a single fixed waveform shape. This behavior matches the design goal of the model, which is to describe the shared cardiac cycle while remaining sensitive to how each sensor renders it.

\subsection{Delay Alignment Visualization}
\begin{figure}[!tbp]
\centering
\scriptsize
\includegraphics[width=0.7\columnwidth]{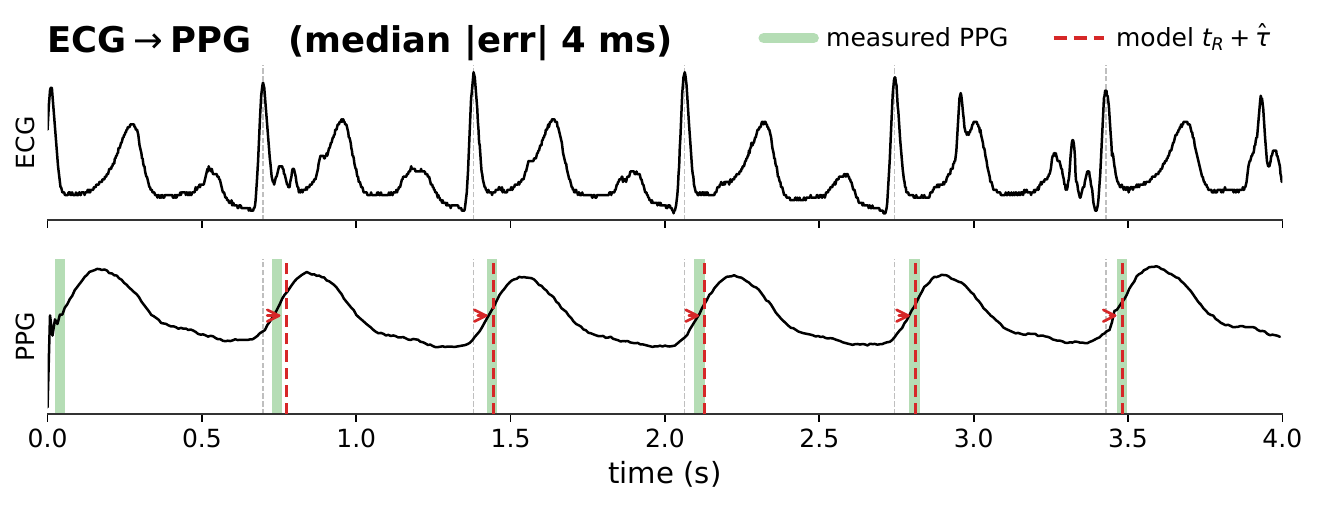} \\[-1mm]
(a) ECG to PPG (VitalDB) \\[3mm]
\includegraphics[width=0.7\columnwidth]{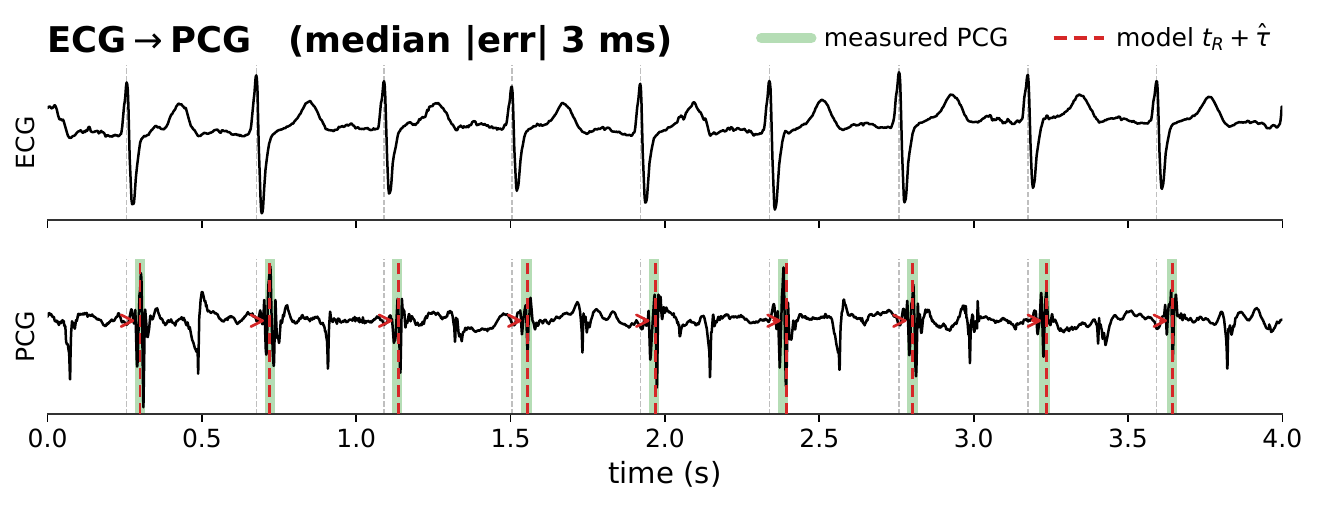} \\[-1mm]
(b) ECG to PCG (EPHNOGRAM)
\caption{Learned delay alignment on paired recordings. The top panel shows ECG to PPG alignment on VitalDB and the bottom panel shows ECG to PCG alignment on EPHNOGRAM. For each ECG R-peak $t_R$, the predicted event time $t_R+\hat{\tau}$ (red dashed) is overlaid on the target signal alongside the actual event (green), namely the systolic upstroke for PPG and the first heart sound $S_1$ for PCG. The learned delay tracks the physiological offset between modalities.}
\label{fig:delay_align}
\end{figure}

To check that the delay aligner captures physiological timing rather than an arbitrary offset, we visualize its output on paired recordings. For each R-peak $t_R$ detected in the ECG, we mark the predicted event time $t_R+\hat{\tau}$ on the target signal and compare it with the true cardiac event in that modality. As shown in Figure~\ref{fig:delay_align}, the predicted time lines up with the systolic upstroke in the PPG and with the first heart sound $S_1$ in the PCG, and it does so consistently from one beat to the next. This close agreement indicates that the learned delay reflects the true electromechanical and pulse transit timing between modalities rather than a value that merely minimizes the training loss. It also gives a direct and interpretable check on the central mechanism of \SysName, since the alignment can be read against known cardiac landmarks.

\subsection{Task-Relevant Structure After Cross-Modal Pretraining}
The shared space analysis showed that cross-modal training makes the pooled cardiac code invariant to the sensing modality, since the modality silhouette falls from $0.121$ to $-0.006$. Here we show a complementary effect within a single modality, namely that the same training makes the PPG representation more discriminative for downstream classes. In both of the following figures we compare PPG features from the Stage~I and Stage~II checkpoints, so the comparison isolates the contribution of cross-modal pretraining rather than comparing against baselines.

\begin{figure}[!tbp]
\centering
\includegraphics[width=0.7\columnwidth]{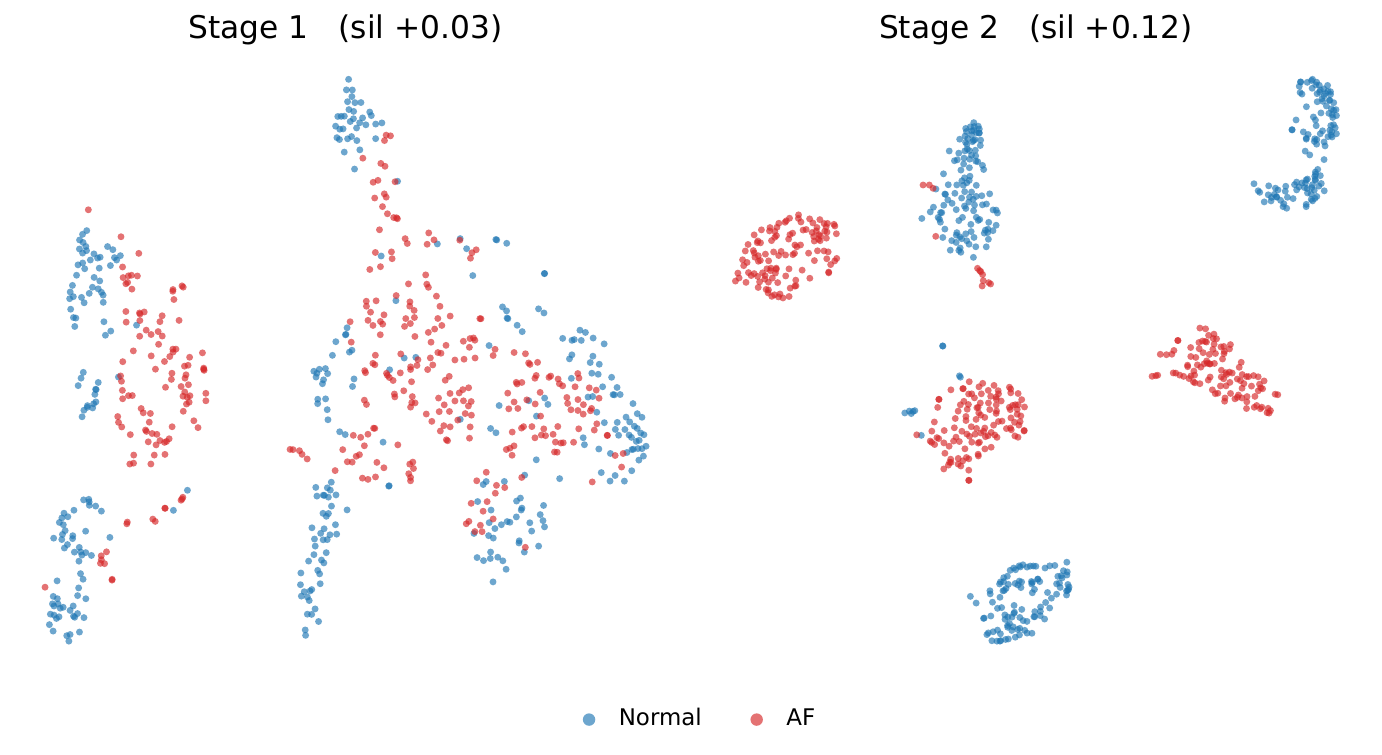}
\caption{t-SNE of PPG features on the binary atrial fibrillation task (MIMICPerform-AF) from the Stage~I (left) and Stage~II (right) checkpoints, colored by class. The class silhouette rises from $0.03$ to $0.12$.}
\label{fig:tsne_af}
\end{figure}

Figure~\ref{fig:tsne_af} shows the binary atrial fibrillation task. After Stage~I the two classes overlap, and after Stage~II they form clearer clusters, with the class silhouette rising from $0.03$ to $0.12$. This shows that pulling the modalities into a shared space does not blur the class structure that a downstream classifier needs, but instead makes it easier to separate.

\begin{figure}[!tbp]
\centering
\includegraphics[width=0.7\columnwidth]{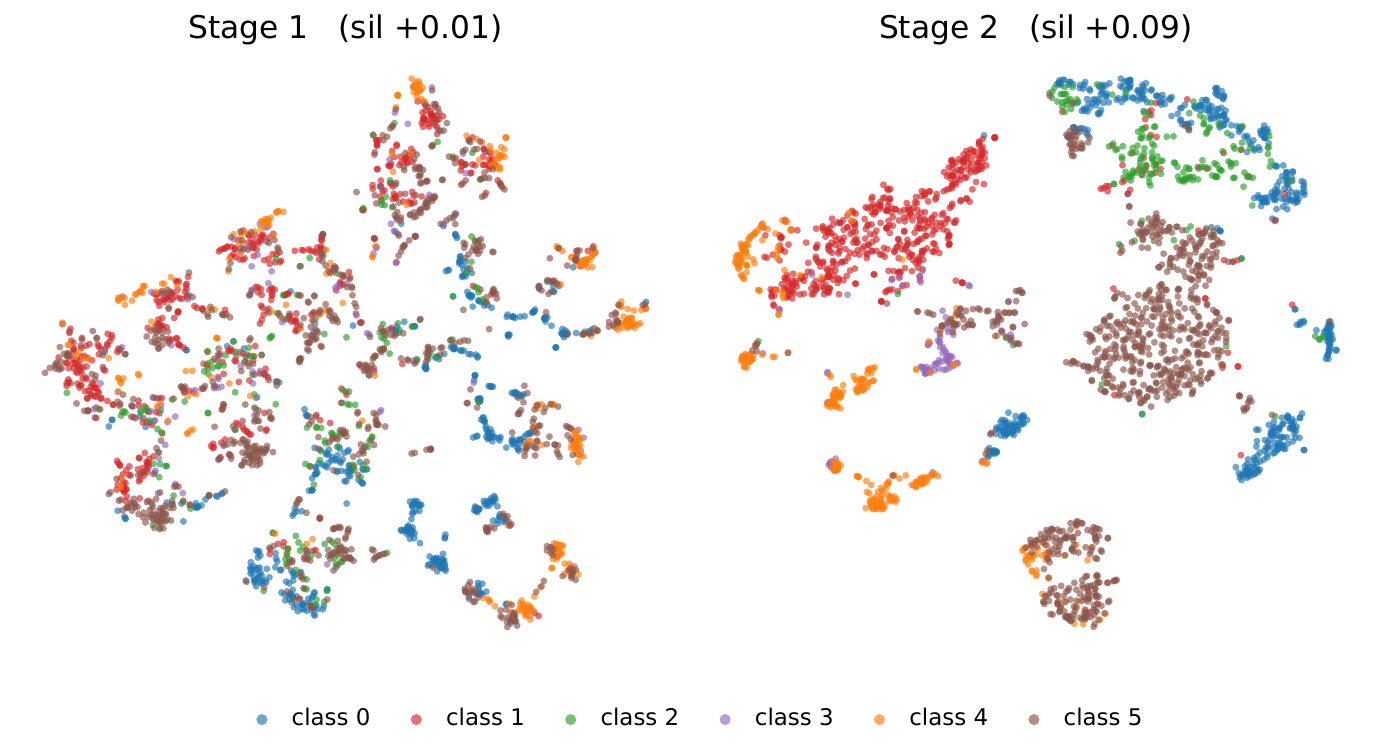}
\caption{t-SNE of PPG features on the six-class arrhythmia task from the Stage~I (left) and Stage~II (right) checkpoints, colored by class. The class silhouette rises from $0.01$ to $0.09$.}
\label{fig:tsne_arr}
\end{figure}

Figure~\ref{fig:tsne_arr} shows the six-class arrhythmia task, where the effect is stronger. Stage~I features are largely intermixed with a silhouette of $0.01$, whereas after Stage~II the classes organize into coherent groups with a silhouette of $0.09$. Because absolute silhouette values are small for t-SNE embeddings of high dimensional features, we read these numbers as relative improvements rather than measures of separability. The modality analysis and the class analysis are computed over different labelings, so cross-modal training is expected to lower the modality silhouette while raising the class silhouette, and together the two figures indicate that the training removes sensor identity from the shared space while sharpening task relevant structure within each modality.

\enlargethispage{3\baselineskip}
\subsection{Limitations}
\SysName has several limitations that also point to future work. First, the paired and trimodal corpora that drive cross modal alignment are much smaller than the unimodal corpora, so the amount of synchronized supervision is limited and the delay aligner is trained on comparatively few clean beats. Second, the PCG pretraining data is the smallest of the three modalities, which places more weight on cross modal transfer for acoustic tasks and may limit how far the acoustic representation can be pushed on its own. Third, on ECG the trimodal model is within noise of the strongest bimodal variant, so the benefit of adding a third modality is clearest for PPG and PCG rather than for ECG. Fourth, the delay aligner assumes that a reliable reference event can be detected on the source signal, so very noisy recordings where beats cannot be located fall back to unsupervised alignment. Finally, all results use a frozen encoder with linear probing, and we leave full fine tuning and larger encoder sizes to future work. We view these limitations as natural next steps rather than obstacles, since each one is tied to data availability or evaluation budget rather than to the core formulation.

\end{document}